\documentclass[journal]{IEEEtran}
\usepackage{cite}

   \usepackage{graphicx}
   \graphicspath{{../eps/}}
\usepackage{epstopdf}

\usepackage{tikz}
\usetikzlibrary{arrows} 
\tikzset{
    >=stealth',
    image/.style={
           rectangle,
		   fill=yellow!10,
           draw=black, very thick,
           text width=10em,
           minimum height=2em,
           text centered},
    process/.style={
           rectangle,
           rounded corners,
		   fill=blue!10,
           draw=black, very thick,
           text width=8em,
           minimum height=2em,
           text centered},
    object/.style={
           circle,
		   fill=yellow!10,
           draw=black, very thick,
           text width=2em,
           minimum height=1.5em,
           text centered},
    pil/.style={
           ->,
           thick,
           shorten <=2pt,
           shorten >=2pt,}
}

\usepackage{color}
\usepackage{colortbl}

\usepackage[cmex10]{amsmath}
\usepackage{amsfonts}
\usepackage{mathtools}
\usepackage{dsfont}
\usepackage{siunitx}
\usepackage{soul}
\usepackage{hyperref}

\DeclareMathOperator*{\argmin}{arg\,min\,}
\DeclareMathOperator{\Tr}{Tr}
\newcommand{\B}{\mathbf{B}} 
\newcommand{\Z}{\mathbf{Z}} 
\newcommand{\Y}{\mathbf{Y}} 
\newcommand{\A}{\mathbf{A}} 
\newcommand{\X}{\mathbf{X}} 
\newcommand{\E}{\mathbf{E}} 
\newcommand{\M}{\mathbf{M}} 
\newcommand{\D}{\mathbf{D}} 
\newcommand{\R}{\mathbf{R}} 
\newcommand{\N}{\mathbf{N}} 
\newcommand{\V}{\mathbf{V}} 
\newcommand{\eye}{\mathbf{I}} 
\newcommand{\Hm}{\mathbf{H}} 
\definecolor{light}{gray}{.9}

\usepackage{algo}
\usepackage{algorithmic}

\usepackage{array}

\ifCLASSOPTIONcompsoc
  \usepackage[caption=false,font=normalsize,labelfont=sf,textfont=sf]{subfig}
\else
  \usepackage[caption=false,font=footnotesize]{subfig}
\fi
\begin{document}
%
\title{A convex formulation for hyperspectral image superresolution via subspace-based regularization}
%
%
%

\author{Miguel Sim\~{o}es, Jos\'{e} Bioucas-Dias,~\IEEEmembership{Member,~IEEE,} Luis B. Almeida, Jocelyn Chanussot,~\IEEEmembership{Fellow,~IEEE,}
\thanks{Miguel Sim\~{o}es is with Instituto de Telecomunica\c{c}\~{o}es, Instituto
Superior T\'{e}cnico, Universidade de Lisboa, Lisbon, Portugal and GIPSA-Lab, Universit\'{e} de Grenoble, France.}
\thanks{Jos\'{e} Bioucas-Dias and Luis B. Almeida are with Instituto de Telecomunica\c{c}\~{o}es, Instituto
Superior T\'{e}cnico, Universidade de Lisboa, Lisbon, Portugal.}
\thanks{Jocelyn Chanussot is with GIPSA-Lab, Grenoble Institute of Technology, France and with the Faculty of Electrical and Computer Engineering, University of Iceland.}
\thanks{This work was partially supported by the Funda\c{c}\~{a}o para a Ci\^{e}ncia e Tecnologia, Portuguese Ministry of Science and Higher Education, projects PEst-OE/EEI/0008/2013,  PTDC/EEI-PRO/1470/2012, and grant SFRH/BD/87693/2012.}}
\maketitle

\begin{abstract}
Hyperspectral remote sensing images (HSIs) usually have high spectral resolution and low spatial resolution. Conversely, multispectral images (MSIs) usually have low spectral and high spatial resolutions. The problem of inferring images which combine the high spectral and high spatial resolutions of HSIs and MSIs, respectively, is a data fusion problem that has been the focus of recent active research due to the increasing availability of HSIs and MSIs retrieved from the same geographical area.

We formulate this problem as the minimization of a convex objective function containing two quadratic data-fitting terms and an edge-preserving regularizer. The data-fitting terms account for blur, different resolutions, and additive noise. The regularizer, a form of vector Total Variation, promotes piecewise-smooth solutions with discontinuities aligned across the hyperspectral bands.

The downsampling operator accounting for the different spatial resolutions, the non-quadratic and non-smooth nature of the regularizer, and the very large size of the HSI to be estimated lead to a hard optimization problem. We deal with these difficulties by exploiting the fact that HSIs generally ``live'' in a low-dimensional subspace and by tailoring the Split Augmented Lagrangian Shrinkage Algorithm (SALSA), which is an instance of the Alternating Direction Method of Multipliers (ADMM), to this optimization problem, by means of a convenient variable splitting. The spatial blur and the spectral linear operators linked, respectively, with the HSI and MSI acquisition processes are also estimated, and we obtain an effective algorithm that outperforms the state-of-the-art, as illustrated in a series of experiments with simulated and real-life data.
\end{abstract}


%

\section{Introduction} \label{sec:introduction}
\IEEEPARstart{I}{mages} are an efficient way to describe and store visual information about our world. This work will deal with a special kind of them, the so-called spectral images. A spectral image, or \emph{data cube}, is a set of 2D images, also termed \emph{bands}, representing the reflectance or radiance of a scene in different parts of the electromagnetic (EM) spectrum. They find applications in the fields of remote sensing (agriculture, mineralogy, etc.), astronomy, and biomedicine, for example~\cite{Bioucas2013}. Our focus will be on the remote sensing field, where spectral images are typically generated from air- or spaceborne sensors.

In this context, it is common to distinguish between hyperspectral and multispectral images (HSIs and MSIs, respectively). The difference is application-dependent, but HSIs typically have high spectral resolution in the visible, near-infrared, and shortwave infrared spectral ranges~\cite{Bioucas2013}. As a result of this high resolution, HSIs  have a large number of bands, each one corresponding to a somewhat narrow part of the EM spectrum. For example, the Hyperion Imaging Spectrometer has about 200 spectral bands, each covering \SI{10}{\nano\metre} of the spectrum, with a spatial resolution of \SI{30}{\metre}~\cite{Middleton2003}.\footnote{More information at \url{http://eo1.usgs.gov/sensors/hyperion}.} On the other hand, MSIs generally offer a higher spatial resolution, but each band covers a larger range of the spectrum, resulting in a much smaller number of bands. For example, the IKONOS satellite collects multispectral images covering four bands (blue, green, red and near-infrared) with a spatial resolution of \SI{3.2}{\metre}~\cite{Kramer1994}.\footnote{More information at \url{http://www.digitalglobe.com/sites/default/files/DG_IKONOS_DS.pdf}.} In other words, HSIs have comparatively high spectral and low spatial resolutions, while MSIs have low spectral and high spatial resolutions. 

It is of interest to fuse the information from these two data sources, to synthesize images with simultaneously high spectral and high spatial resolutions. A related problem that has been extensively studied is pansharpening, which addresses the fusion of multispectral and panchromatic images, the latter of which are single-band images usually covering the visible and the  near-infrared spectral ranges~\cite{Alparone2007, Thomas2008, Amro2011}. Panchromatic images (PANs) typically have a spatial resolution that is even higher than the one of MSIs. The HSI-MSI fusion problem is significantly more difficult to solve than pansharpening, owing to three factors: a) although both are ill-posed, there is a much larger number of variables to estimate in HSI-MSI fusion, b) the hyperspectral data typically have a large dimensionality, which can act computationally more as a crutch than an asset, and c) often, the spectral range covered by the HSI is significantly larger than the one covered by the MSI, and, therefore, many bands of the HSI are not included in any band of the MSI.

Recently, some techniques dedicated to the fusion of HSIs and MSIs have been proposed. A common trend is to associate this problem with the linear spectral unmixing one, which assumes that the underlying data can be described by a mixture of a relatively small number of ``pure'' spectral signatures, corresponding to the materials that are present in the scene~\cite{Bioucas2013, Greer2012}. Since both HSIs and MSIs capture the same scene, the underlying materials (the so-called \emph{endmembers}) should be the same. Therefore, a spectral dictionary extracted from one of the images should also be able to explain the other one. Due to the high spectral resolution of the HSIs, the dictionary is extracted from these data, and is then used to reconstruct the multispectral data via sparse regression. The estimate of the original high resolution HSI is then obtained from the regression coefficients and from the dictionary. This technique was introduced in~\cite{Charles2011} for HSIs, but there are older works exploiting similar ideas for MSIs~\cite{Zurita2008}. For example, Kawakami \textit{et al.}~\cite{Kawakami2011a} fused hyperspectral images with images from RGB cameras, starting by estimating the endmember mixing matrix from the hyperspectral data through a $\ell_1$-minimization problem, solved via a non-smooth Gauss-Newton algorithm. The endmember matrix, jointly with the spectral responses of the RGB sensor, was then used as a basis to reconstruct the RGB image, by formulating an optimization problem that imposed sparsity. In~\cite{Huang2013}, Huang \textit{et al.} unmixed the hyperspectral data via the K-SVD algorithm, and reconstructed the MSI using orthogonal matching pursuit to induce sparsity. The method was tested with Landsat/ETM+ and Aqua/Modis images. Song \textit{et al.}~\cite{Song2013} first learned two dictionaries from the two different data, and then used a dictionary-pair learning method to establish the correspondence between them. Again, their method was tested using Landsat/ETM+ with Aqua/Modis data, but only taking into account the spectrally overlapping bands. A similar and older technique is the one from Yokoya \textit{et al.}, which alternately unmixes both sources of data to find the signatures and the abundances of the endmembers~\cite{Yokoya2012}.

A different framework was proposed by Hardie \textit{et al.} in~\cite{Hardie2004}, in which a fully Bayesian approach was followed, by imposing prior distributions on the problem. This work was the foundation for other works: Zhang \textit{et al.} introduced a method which works in the wavelet domain~\cite{Zhang2009}, and later published an expectation-–maximization algorithm to maximize the posterior distribution~\cite{Zhang2012}. Wei \textit{et al.} used a Hamilton Monte Carlo algorithm to deal with the high-dimensional space of the posterior distribution~\cite{Wei2014}. In~\cite{Chen2011}, Chen \textit{et al.} introduced a method that treats image registration and image fusion as a joint process. The fusion of HSIs with just the panchromatic band is a different, but related, problem~\cite{Khan2009, Garzelli2010, Licciardi2012}. Using only the hyperspectral image, different authors~\cite{Akgun2005, Zhao2011} treated this problem as a simple superresolution one. 

\subsection{Contributions}

This work is built around the standard linear inverse problem model for HSIs and MSIs. This model is used to formulate data fusion as a convex optimization problem. We use a form of vector Total Variation-based regularization~\cite{Bresson2008}, taking into account both the spatial and the spectral characteristics of the data. In order to perform the optimization, we follow an Alternating Direction Method of Multipliers (ADMM) approach by using the Split Augmented Lagrangian Shrinkage Algorithm (SALSA)~\cite{Afonso2011}, and we explore the inherent redundancy of the images with data reduction techniques, to formulate the problem in a computationally efficient way. This method, which we term \emph{HySure}, for Hyperspectral Superresolution, allows us to fuse hyperspectral data with either multispectral or panchromatic images.

In the literature, the HSI-MSI fusion problem is very often dealt with as a non-blind one, in the sense that the spatial and spectral responses of the sensors are assumed to be known (see~\cite{Yokoya2012, Zhang2009, Zhang2012, Wei2014}, for example). In practice, however, the information that is available about these responses is often scarce and/or somewhat inaccurate. In this work, we take a blind approach, assuming that these responses are unknown, and we formulate another convex problem to estimate them, making only minimal assumptions: we assume that the spatial response has limited support and that both responses are relatively smooth. The estimate of the spectral response can be improved by using information on the correspondence between bands from the two images, if that information is available---it is often easily obtained from data on the spectral coverage of the various bands from the two sensors.

This work extends~\cite{Simoes2014} in several different directions: it details the optimization process more clearly, it establishes the framework used to estimate the spatial and spectral responses of the sensors, and it presents a number of new experimental results.

\subsection{Outline}

The remainder of this work is organized as follows. Section~\ref{sec:method} describes the data fusion method, including the proposed model and the formulation of the optimization problem. The approach followed to perform the optimization is presented in Section~\ref{sec:admm}. Section~\ref{sec:estimate} deals with the estimation of the sensors' spatial and spectral responses. Section~\ref{sec:exp} presents experimental results. Section~\ref{sec:conclusions} concludes.

%
%

\section{Data Fusion Method} \label{sec:method}

\subsection{Observation Model}

\label{subsec:obser_model}

Multispectral and hyperspectral images can be thought of as three-dimensional arrays or tensors, often called data cubes. However, for notational convenience, the representation followed in this work will consider HSIs and MSIs to be two-dimensional matrices, where each line corresponds to a spectral band, containing the lexicographically ordered pixels of that band. We use bold lowercase to denote vectors (e.g., $\mathbf{x}$, $\mathbf{y}$) and bold uppercase to denote matrices (e.g., $\mathbf{H}$, $\mathbf{M}$).

Let the matrix representing the observed hyperspectral data be $\Y_h \in \mathbb{R}^{L_h \times n_h}$, with $L_h$ bands and spatial dimension $n_h$, and let $\Y_m \in \mathbb{R}^{L_m \times n_m}$ denote the observed multispectral data, with $L_m<L_h$ bands and spatial dimension $n_m>n_h$. Matrix $\Z \in \mathbb{R}^{L_h \times n_m}$ denotes the high spatial and spectral resolution data to be estimated. 

With this representation, we model the hyperspectral measurements as
\begin{equation} \label{eq:model_y_h}
\Y_h = \Z \B \M + \N_h
\end{equation}
where matrix $\B \in \mathbb{R}^{n_m \times n_m}$ is a spatial blurring matrix representing the hyperspectral sensor's point spread function in the spatial resolution of $\Z$; it is assumed to be band-independent and to be under circular boundary conditions. These two assumptions are made for simplicity. When dealing with non-blind data fusion, allowing the blur to vary across bands would not change the complexity of the algorithm. In the blind case, the increase in complexity would be relatively small. Regarding the boundary conditions, assuming them to be periodic has two main advantages. First, it allows to use Fast Fourier Transforms (FFTs) to compute convolutions. Second, matrix inversion, usually a costly operation, is easily performed, again through the use of FFTs, under certain conditions that are met in this case. 
Although periodic boundary conditions are not totally realistic, we experimentally found that they do not lead to any significant artifacts in the fused image, while allowing a dramatic reduction in the amount of computation. A technique based on ADMM that makes no assumptions about the boundaries has been proposed in~\cite{Almeida2013a, Matakos2013}, but we did not find the corresponding increase in complexity justified for the images we have worked on.

Matrix $\M \in \mathbb{R}^{n_m \times n_h}$, whose columns are a subset of the columns of the identity matrix, accounts for a uniform subsampling of the image, to yield the lower spatial resolution of the hyperspectral image. $\N_h$ represents independent and identically distributed (i.i.d.) noise. The assumption that the noise is identically distributed across bands is also made for simplicity. Accommodating statistically independent noise across bands and pixels, but with band-dependent variance, would be straightforward.

We model the multispectral measurements as
\begin{equation} \label{eq:model_y_m}
\Y_m = \R \Z + \N_m,
\end{equation}
where $\R \in \mathbb{R}^{L_m \times L_h}$ holds in its rows the spectral responses of the multispectral instrument, one per multispectral band, and $\N_m$ represents i.i.d. noise.

In this work, matrices $\B$ and $\R$ are estimated from the data, by formulating a quadratic optimization problem. Section~\ref{sec:estimate} will address that topic.

\subsection{Dimensionality reduction}
\label{sec:dimreduction}

Hyperspectral data normally have a large correlation between bands: the spectral vectors, of size $L_h$, usually ``live'' in a subspace of dimension much lower than $L_h$~\cite{Bioucas-Dias2012, Cawse2013}. Therefore, we can write
\begin{equation} \label{eq:Z}
\Z = \E \X,
\end{equation}
where $\E \in \mathbb{R}^{L_h \times L_s}$ is a matrix whose $L_s$ columns span the same subspace as the columns of $\Z$, and $\X \in \mathbb{R}^{L_s \times n_m}$ are the representation coefficients. Small values of $L_s$, i.e., $L_s \ll L_h$, translate into a description of the data in a relatively low-dimensional space. 

This dimensionality reduction has two advantages. One is that it is computationally more efficient to work in a lower dimensional space than in the original space of $\Z$, making algorithms which use these representations comparatively fast. The other advantage is that, since the number of variables to be estimated is significantly reduced, the estimates will normally be more accurate than if we worked in the original dimensionality. As an illustration of the amount of reduction that is possible, assume that the hyperspectral image has 200 bands. With $L_s=10$, which is a typical value, only $5 \%$ of the number of original variables need to be inferred.

Different approaches can be followed to factorize matrix $\Z$, and two of them will be briefly mentioned here. One is to take into account the physical process that gave origin to $\Y_h$. In the linear unmixing approach~\cite{Bioucas-Dias2012}, it is assumed that the spectral response of each pixel is a linear combination of the pure spectral signatures of the underlying endmembers. In this case, $\E$ would be the spectral signature matrix obtained from $\Y_h$, and $\X$ would represent the abundance fractions of the endmembers for every pixel of $\Z$. There are numerous algorithms in the literature that address the unmixing problem (for example, Vertex Component Analysis -- VCA~\cite{Nascimento2005}). Several of the methods discussed in Section~\ref{sec:introduction} use the linear mixing model.

Another approach is to use Singular Value Decomposition (SVD) to obtain the factorization $\Y_h = \mathbf{U} \mathbf{\Sigma} \mathbf{V^T}$, where $\mathbf{U}$ and $\mathbf{V}$ are orthogonal matrices and $\mathbf{\Sigma}$ is a rectangular diagonal matrix containing the singular values, which are assumed to be in non-increasing order. Denote by $\hat{\mathbf{\Sigma}}$, $\hat{\mathbf{U}}$ and $\hat{\mathbf{V}}$, respectively, the truncated matrices obtained by discarding the rows and columns with the smallest singular values from $\mathbf{\Sigma}$ and the corresponding columns of $\mathbf{U}$ and $\mathbf{V}$. A low-dimensional approximation of $\Y_h$ is given by $\hat{\mathbf{U}} \hat{\mathbf{\Sigma}} \hat{\mathbf{V}}^T$. In this approach, we make $\E = \hat{\mathbf{U}}$. Due to the low intrinsic dimensionality of the hyperspectral data, most of the singular values are rather small, allowing a very significant dimensionality reduction while retaining a rather faithful approximation of $\Y_h$. If $\N_h=\mathbf{0}$ and all discarded singular values are zero, this representation spans the true signal subspace. If the former condition on $\N_h$ is not obeyed but $\N_h$ is i.i.d., this representation corresponds to the maximum likelihood estimate of that subspace. However, if the noise is non-i.i.d., the estimation of the subspace is more complex; see, for example, \cite{Bioucas-Dias2008} for details, and for algorithms oriented to subspace estimation in hyperspectral applications.

With any of these two factorizations, we replace Eq.~\eqref{eq:model_y_h} with
\begin{equation} \label{eq:model_y_h_reduced}
\Y_h = \E \X \B \M + \N_h,
\end{equation}
where the error due to the dimensionality reduction has been incorporated into $\N_h$.

Remote sensing images often are somewhat noisy. The use of truncated SVD is also a very common approach to perform denoising, a topic that we shall address in Section~\ref{sec:exp}.

\subsection{Regularization}

The problem that we are trying to solve is strongly ill-posed, and therefore needs adequate regularization. The regularizer that we use is given by
\begin{equation}
\varphi\big(\X \D_h, \X \D_v \big) \; {\buildrel\rm def\over=} \; \sum_{j=1}^{n_m} \sqrt{\sum_{i=1}^{L_s} \Big\{\big[(\X \D_h)_{ij}\big]^2 + \big[(\X \D_v)_{ij}\big]^2\Big\}},
\end{equation}
where $(\mathbf{A})_{ij}$ denotes the element in the $i$th row and $j$th column of matrix $\mathbf{A}$, and the products by matrices $\D_h$ and $\D_v$ compute the horizontal and vertical discrete differences of an image, respectively, with periodic boundary conditions. This regularizer is a form of vector Total Variation (VTV)~\cite{Bresson2008}. Its purpose is to impose sparsity in the distribution of the absolute gradient of an image, meaning that transitions between the pixels of an image should be smooth in the spatial dimension, except for a small number of them, which should coincide with details such as edges. Total Variation was proposed for the first time in~\cite{Rudin1992} and is extensively used in image restoration~\cite{Chambolle2004, Wang2008, Tao2009, Afonso2011, Almeida2013a, Sroubek2012, Xu2010, campisi2007}. It has two different discrete formulations, the anisotropic and isotropic ones \cite{Beck2009}; in this work, we use the isotropic formulation. In \cite{Zhao2013}, Zhao \textit{et al.} proposed an isotropic TV scheme for hyperspectral image deblurring in a band-by-band manner. This means that each band was regularized independently from the other ones. This approach has a shortcoming: it does not take into account that edges and other details normally have the same locations in most bands. The vector form of the regularizer, which we use in this work, promotes solutions in which edges and other details are aligned among the different bands. VTV has previously been used in a pansharpening application~\cite{He2014} and in the denoising of hyperspectral images~\cite{Yuan2012}. 

We apply the regularizer to the reduced-dimensionality data  $\X$, and not to $\Z$ itself. This is indeed reasonable, since the subspace spanned by $\E$ is the same as the one where $\Z$ resides (or an approximation, when using truncated SVD), and by regularizing $\X$ we are indirectly regularizing $\Z$.

\subsection{Optimization problem}

Let $\| \X \|_F \; {\buildrel\rm def\over=} \; \sqrt{\Tr(\X \X^T)}$ denote the Frobenius norm of $\X$, and $(\cdot)^T$ denote the transposition operator. We can now formulate an optimization problem based on our model with the proposed regularizer: 
\begin{align} \label{eq:optimizationproblem}
& \underset{\X}{\text{minimize}}
& & \frac{1}{2}\Big\|\Y_h - \E\X\B\M \Big\|_F^2 + \frac{\lambda_{m}}{2}\Big\|\Y_m - \R\E\X \Big\|_F^2 \nonumber \\
& & & \quad + \lambda_{\varphi} \varphi \big(\X \D_h, \X \D_v \big).
\end{align}
The first two terms are data-fitting terms, imposing that the estimated image should be able to explain the observed data according to the model defined in~(\ref{eq:model_y_h_reduced}) and~(\ref{eq:model_y_m}). The last term is the regularizer. The parameters $\lambda_m$ and $\lambda_{\varphi}$ control the relative importances of the various terms. We shall discuss the selection of these parameters in Section~\ref{sec:implementation_details}.

Problem~\eqref{eq:optimizationproblem} is convex, but is rather hard to solve, due to the nature of the regularizer, which is non-quadratic and non-smooth. Additional difficulties are raised by the large size of $\X$ (the variable to be estimated) and by the presence of the downsampling operator $\M$ in one of the quadratic terms, preventing a direct use of the Fourier transform in optimizations involving this term. We deal with these difficulties by {using \mbox{SALSA~\cite{Afonso2011}}}. An alternative approach would consist in employing a primal-dual method~\cite{Chambolle2011, Condat2013}. Unlike our approach, primal-dual methods do not require the solution of linear systems of equations on each iteration. However, since the system matrix in our problem is diagonalizable using light computations, SALSA yields much faster algorithms than those based on primal-dual methods, according to our experience. The next section and the Appendix describe the details of the optimization method.

\section{Optimization method} \label{sec:admm}

ADMM involves the introduction of auxiliary variables into the optimization problem, through the so-called variable splitting technique. We split the original optimization variable $\X$ into a total of five variables: one which we still call $\X$, and four auxiliary variables, $\V_1$ to $\V_4$. The optimization problem becomes
\begin{align} \label{eq:constraint}
& \underset{\X, \V_1, \V_2, \V_3, \V_4}{\text{{minimize}}}
& & \frac{1}{2}\Big\|\Y_h - \E\V_1\M \Big\|_F^2 + \frac{\lambda_{m}}{2}\Big\|\Y_m - \R\E\V_2 \Big\|_F^2 \nonumber \\
& & & \quad + \lambda_{\varphi} \varphi \big(\V_3, \V_4 \big) \nonumber\\
& \text{subject to}
& & \V_1 = \X \B,\\
& & & \V_2 = \X, \nonumber\\
& & & \V_3 = \X\D_h, \nonumber \\
& & & \V_4 = \X\D_v. \nonumber
\end{align}
For notational simplicity, we define the matrices $\V$ and $\Hm$,
\[
\V \; {\buildrel\rm def\over=} \; 
\begin{bmatrix}
\V_1^T \\ 
\V_2^T \\ 
\V_3^T \\ 
\V_4^T
\end{bmatrix},
\quad \Hm \; {\buildrel\rm def\over=} \;
\begin{bmatrix}
\B^T \\
\eye \\
\D_h^T \\
\D_v^T
\end{bmatrix},
\] 
and the cost function as follows:
\[
\begin{aligned}
f\big(\V \big)\; {\buildrel\rm def\over=} \; & \frac{1}{2}\Big\|\Y_h - \E\V_1\M \Big\|_F^2 + \frac{\lambda_{m}}{2}\Big\|\Y_m - \R\E\V_2 \Big\|_F^2 \\
& \quad + \lambda_{\varphi} \varphi \big(\V_3, \V_4 \big).
\end{aligned}
\] 
We can express~(\ref{eq:constraint}) as 
\begin{equation} \label{eq:simplicity}
\begin{aligned}
& \underset{\X, \V}{\text{{minimize}}}
& & f\big(\V \big)\\\
& \text{subject to}
& & \V = \Hm \X^T.		
\end{aligned}
\end{equation}
This problem has the following augmented Lagrangian~\cite{Nocedal2006}
\begin{equation} \label{eq:al}
\begin{aligned}
\mathcal{L}\big(\X, &\V, \A \big) = f\big(\V \big) + \frac{\mu}{2}\Big\|\Hm \X^T - \V - \A\Big\|_F^2,\\
\end{aligned}
\end{equation}
where $\A$ is the so-called scaled dual variable \cite{Boyd2011}, and $\mu$ is a positive constant, called penalty parameter. We are now ready to apply the ADMM method, which yields the algorithm shown in Fig.~\ref{alg:salsa}. As we can see, SALSA solves the original, complex optimization problem through an iteration on a set of much simpler problems. The constraints are taken into account, in an approximate way, by minimizing the augmented Lagrangian of the problem relative to the auxiliary variables.

\begin{figure}[tbh!]
\begin{center}
\colorbox{light}{\parbox{1.0\columnwidth}{
\begin{algorithmic}
\REQUIRE data: $\Y_h$, $\Y_m$; regularization parameters: $\lambda_m$, $\lambda_{\varphi}$; penalty parameter: $\mu$; matrices $\R$, $\B$ and $\E$; initializations: $\V^{(0)}$ and $\A^{(0)}$
\STATE $k \coloneqq 0$
\REPEAT
\STATE $\mathbf{X}^{(k+1)} \in \underset{\mathbf{X}}\argmin \enspace \mathcal{L}\Big(\mathbf{X}, \mathbf{V}^{(k)}, \mathbf{A}^{(k)} \Big)$
\STATE $\mathbf{V}^{(k+1)} \in \underset{\mathbf{V}}\argmin \enspace \mathcal{L}\Big(\mathbf{X}^{(k+1)}, \mathbf{V}, \mathbf{A}^{(k)}\Big)$
\STATE $\mathbf{A}^{(k+1)} \coloneqq \mathbf{A}^{(k)} - \Big(\Hm {\X^{(k+1)}}^T - \V^{(k+1)} \Big)$
\STATE $k \coloneqq k+1$
\UNTIL{stopping criterion is satisfied.}
\end{algorithmic} 
}}
\caption{Pseudocode for the HySure algorithm. For details, see the Appendix.}
\label{alg:salsa}
\end{center}
\end{figure}

The minimization with respect to $\X$ is a quadratic problem with a block cyclic system matrix, which can be efficiently solved by means of the Fast Fourier Transform (FFT). Minimizing with respect to the auxiliary variables is done by solving three different problems, whose solutions correspond to three Moreau proximity operators~\cite{Combettes2006}. The minimization with respect to $\V_1$ is a quadratic problem which is efficiently solved via FFTs, and the minimization relative to $\V_2$ is also quadratic; these two problems involve matrix inverses which can be computed in advance. Finally, the minimization with respect to $\V_3$ and $\V_4$ corresponds to a pixel-wise \textit{vector soft-thresholding} operation.

The details of the optimization, as well as an analysis of the algorithm's complexity, are presented in the Appendix. The number of splitting variables could have been reduced, by eliminating $\V_1$, for example. This could have been done via a scheme similar to the one proposed in~\cite{Zhao2013}, working with Kronecker products. We chose not to do so, since the form of the algorithm that we presented above is simpler to derive, and the computational and memory gains of doing one less splitting did not seem to be very significant. 

The algorithm described above satisfies the conditions for the convergence of SALSA established in~\cite{Afonso2011}, which require matrix $\Hm$ to have full column rank (which is true in our case, due to the presence of identity matrix $\eye$), and function $f(\cdot)$ to be closed, proper, and convex (which is also true, since it is a sum of closed, proper, and convex functions). Under these conditions, and for arbitrary $\mu > 0$, $\V^{(0)}$ and $\A^{(0)}$, if problem~(\ref{eq:simplicity}) has a solution $\X^*$, then the sequence $\{\X^{(k)}\}$ will converge to $\X^*$; if a solution does not exist, then at least one of the sequences $\{\V^{(k)}\}$ or $\{\A^{(k)}\}$ will diverge. The actual value of the penalty parameter $\mu$ is not important as a condition for convergence, but can have a strong influence on the convergence speed of the algorithm. The choice of $\mu$ is discussed in Section~\ref{sec:implementation_details}.

\section{Estimating the spatial blur and the spectral response from the data} \label{sec:estimate}

As previously mentioned, matrices $\B$ and $\R$ are estimated from the observed images. The advantages of doing so are threefold. First, as previously mentioned, the available information about the sensors can be rather scarce. Second, it may be hard to precisely adapt that information to the model that is being used for data fusion. Third, there may be discrepancies between the real spatial and spectral responses and the data supplied by the manufacturers. These can be due to several causes, such as atmospheric conditions, postprocessing artifacts, and even the variability within the observed scene~\cite{Wang2010}.

As already mentioned, in~\cite{Zhang2009, Zhang2012}, Zhang \textit{et al.} assumed the spatial response to be known. However, they also suggested using Gaussian blurs with different variances as spatial responses when this was not the case, arguing that their fusion method did not require a strict knowledge of the spatial response of the sensor. In~\cite{Yokoya2013}, Yokoya \textit{et al.} have directly addressed the estimation of responses for the fusion of HSIs and MSIs from the Hyperion and ASTER sensors, respectively, which are aboard two different satellites. Their method estimates both the relative spatial and relative spectral responses of the sensors. The spatial response is assumed to correspond to a Gaussian blur and its variance is estimated by using a template-matching technique. In order to determine the spectral response, the authors use the so-called pre-launch response, with information obtained from measurements performed on the sensors before they were launched into space. The method tries to find a spectral response that is able to describe the observed data and that is close to the pre-launch response. In a different approach, Huang \textit{et al.} estimated the spectral response directly form the data, without requiring \textit{a priori} information~\cite{Huang2013}. 

Recall that, without noise, 
\[
\Y_h = \Z \B \M, \quad \Y_m = \R \Z,
\]
which implies that
\begin{equation} \label{eq:model_B}
\R \Y_h = \Y_m \B \M.
\end{equation}

Taking (\ref{eq:model_B}) into account, we infer $\bf R$ and $\bf B$ by solving the  optimization problem
\begin{equation}
\label{eq:B_R_optim}
\begin{aligned}
\underset{\mathbf{B,R}}{\text{minimize}} &&
\big\|\R \Y_h - \Y_m \B \M \big\|^2 + \lambda_b \phi_B(\mathbf{B}) + \lambda_R \phi_R(\R),
\end{aligned}
\end{equation}
where $\phi_B(\cdot)$ and $\phi_R(\cdot)$ are quadratic regularizers that will be discussed in detail below,
and $\lambda_b,\lambda_R\geq 0$ are the respective regularization parameters. Matrix $\B$, and possibly also matrix $\R$, are subject to some constraints discussed below.

A special consideration needs to be made regarding the estimation of the spectral response. This is due to the fact that, when using the observed data, it is not possible to fully estimate matrix $\R$. The reason for this is that, as discussed in Section~\ref{sec:dimreduction}, the hyperspectral data normally span only a low-dimensional subspace of the full spectral space. Only the component of $\R$ parallel to that subspace can be estimated. This is not a drawback, however, since the component of $\R$ orthogonal to that subspace has essentially no influence on the result of the image fusion. In fact, if we write $\R = \R_{\parallel} + \R_{\perp}$, where ${\bf R}_{\parallel} ={\bf R P}_\parallel$ and ${\bf R}_{\perp} ={\bf R P}_\perp$, and ${\bf P}_\parallel$ and ${\bf P}_\perp$ denote the projection matrices onto the subspaces spanned by the original hyperspectral vectors and onto the subspace orthogonal to it, respectively, we have $\R \Y_h = \R_{\parallel} \Y_h + \R_{\perp} \Y_h = \R_{\parallel} \Y_h$, since $\R_{\perp} \Y_h$ is zero. For the product $\R\Z$, which is involved in the fusion problem, we have $\R\Z \approx \R_{\parallel}\Z$, since $\Z$ will span approximately the same subspace as $\Y_h$, because it corresponds to an image containing the same endmembers.

According to the observation model presented in Section \ref{subsec:obser_model}, matrix $\B$  accounts for a 2D cyclic convolution. In addition, we assume that the convolution kernel  has finite support  contained in
a square window of size $\sqrt{n_b}$, thus containing $n_b$ pixels, centered at the origin.

Let $[\Y_m \B]_{:j}$ denote the $j$th column of $\Y_m \B$,   ${\bf b}\in \mathbb{R}^{n_b}$ denote the columnwise ordering of the convolution kernel, and $\mathbf{P}_j \in \mathbb{R}^{n_m \times n_b}$ denote a matrix which selects from $\Y_m$ a patch such that
\[
  [\Y_m \B]_{:j} = (\Y_m \mathbf{P}_j){\bf b}.
\]
With these  definitions in place, a slight modification of the optimization (\ref{eq:B_R_optim}) is
\begin{equation}
\label{eq:b_R_optim}
\begin{aligned}
& \underset{\mathbf{b,R}}{\text{minimize}}
& & \sum_{j=1}^{n_h} \Big\| \R {\bf Y}_{h,:j} - \mathbf{Y}_{m, j}\mathbf{b} \Big\|^2 \quad \\
& & & \quad \quad \quad +   \lambda_b \phi_b(\mathbf{b}) + \lambda_R \phi_R(\R)\\
& \text{subject to}
& & \mathbf{b}^T\mathbf{1} = 1,
\end{aligned}
\end{equation}
where ${\bf Y}_{h,:j}$ denotes the $j$th column of $\Y_h$, $\mathbf{Y}_{m,j} {\buildrel\rm def\over=} \big[(\Y_m \mathbf{P}_{c_j})\big]\in\mathbb{R}^{L_m\times n_b}$, with $c_j$ denoting the column of $\Y_m$ corresponding to the $j$th column of $\Y_h$, $\phi_b(\mathbf{b}) {\buildrel\rm def\over=} \phi_B(\mathbf{B})$, and  the normalization condition $\mathbf{b}^T\mathbf{1} = 1$ imposes unit DC gain of the blur.

We note that (\ref{eq:b_R_optim}) is a quadratic program with only equality constraints and, therefore, using Lagrange multipliers, its solution can be obtained by solving a linear system of equations. However, even though we have a closed-form solution, because the size of the optimization variables (i.e., $n_b + L_m\times L_h$) is usually of the order of thousands, it may be useful to solve problem~(\ref{eq:b_R_optim}) via alternated minimization with respect to $\mathbf{b}$ and $\R$.

The optimization with respect to $\mathbf{b}$ leads to the following regularized least squares problem:

\begin{equation} \label{eq:est_B}
\begin{aligned}
& \underset{\mathbf{b}}{\text{minimize}}
& & \sum_{j=1}^{n_h} \Big\| \R \Y_{h,:j} - \mathbf{Y}_{m,j}\mathbf{b}\Big\|^2\\
& & & \quad + \lambda_b \Big( \big\|\D_h \mathbf{b} \big\|^2 + \big\|\D_v \mathbf{b} \big\|^2 \Big)\\
& \text{subject to}
& & \mathbf{b}^T\mathbf{1} = 1,
\end{aligned}
\end{equation}

The two last terms of the function being minimized in~\eqref{eq:est_B} correspond to $\phi_b(\cdot)$, which is a noise-removing regularizer that smooths the estimated convolution kernel by promoting that the values of the differences between neighboring pixels be small. As before, $\D_h$ and $\D_v$ compute the horizontal and vertical discrete differences of the convolution kernel, with dimensions adjusted for this particular case.

An approximate solution for \eqref{eq:est_B} is computed by first relaxing the constraint, estimating the filter without the normalization condition, and then normalizing the result to unit DC gain. The solution of the unconstrained problem is given by

\begin{equation}
\label{eq:est_B_sol}
\begin{split}
 \mathbf{b}^* = \Big[ \sum_{j=1}^{n_h} \mathbf{Y}_{m,j}^T \mathbf{Y}_{m,j}  + \lambda_b \big(\D_h^T \D_h + \D_v^T \D_v \big) \Big]^{-1}  \\
  \Big[ \sum_{j=1}^{n_h} \mathbf{Y}_{m,j}^T \R \Y_{h,:j} \Big].
\end{split}
\end{equation}

The support covered by $\mathbf{b}$ is user-specified. We have found, experimentally, that the choice of this support does not have much influence on the blur estimate, as long as it encompasses the support of the actual blur.

Concerning the estimation of $\R$, we use the regularizer $\phi_R(\cdot)$ in order to deal with the indetermination of the orthogonal component, and to reduce estimation noise. In the cases in which there is information about the overlap between bands of the HSI and the MSI, we constrain the elements of $\R$ that correspond to non-overlapping bands to zero.

The estimation of $\R$ can be made independently for each of the MSI bands. Let  $\mathbf{r}_i^T$ denote a row vector containing the $i$th row of $\R$ without the elements that are known to correspond to hyperspectral bands that do not overlap the $i$th multispectral band, and  by $\Y_{h,i}$ denote the matrix $\Y_h$ without the rows corresponding to those same bands. The  optimization of  (\ref{eq:b_R_optim}) is decoupled with respect  to the rows of $\R$ and may be written as

\begin{equation}
\label{eq:est_R}
\underset{\mathbf{r}_i}{\text{minimize}} \quad
\big\|\mathbf{r}_i^T \Y_{h,i} - \Y_{m,i:} \B \M \big\|_F^2 + \lambda_R \big\|\D \mathbf{r}_i \big\|^2,	
\end{equation}
in which $\Y_{m,i:}$ is the $i$th row of $\Y_m$, and the product by $\D$ computes the differences between the elements in $\mathbf{r}_i$ corresponding to contiguous hyperspectral bands. The solution of  (\ref{eq:est_R}) is given by

\begin{equation} \label{eq:est_R_sol}
\mathbf{r}_i^* = \Big[\Y_{h,i} \Y_{h,i}^T + \lambda_R \D^T \D \Big]^{-1} \Y_{h,i} \Big[\Y_{m,i:} \B \M \Big]^T.
\end{equation}

The estimation of each of the matrices $\B$ and $\R$, as presented so far, requires the knowledge of the other matrix. In order to estimate both, and instead of using alternating optimization as proposed before, we adopt an even simpler technique.
We start by estimating $\R$. To do this without knowing $\B$, we first blur both spectral images with a spatial blur that is much stronger than the one produced by $\B$, so that the effect of $\B$ becomes negligible. This, conveniently, also minimizes the effect of possible misregistration between the hyperspectral and multispectral images. Following this, we estimate the spectral response $\R$ using~(\ref{eq:est_R_sol}), setting the kernel of the spatial blur between the strongly blurred multispectral and hyperspectral images to a delta impulse. Finally, we estimate the spatial blur $\B$ using~(\ref{eq:est_B}) on the original (unblurred) images, with the value of $\R$ just found. Fig.~\ref{alg:R_B} summarizes the estimation method. In the tests presented in Section~\ref{sec:exp}, we have used, for the strong spatial blur, an averaging in a square of $9 \times 9$ pixels for the MSI, and a correspondingly smaller averaging for the HSI.

We now discuss the set of solutions of (\ref{eq:b_R_optim}), which is an important issue in our approach to the estimation of $\bf b$ and $\bf R$, closely related to that of identifiability. Given that the objective function is quadratic, a sufficient condition for it to have a unique solution is that its Hessian matrix  be positive definite. Assuming that $\lambda_b, \lambda_R > 0$, the null space associated with the regularization terms  is  the set
\[
{\bf A} {\buildrel\rm def\over=} \{({\bf R,b})\,:\,{\bf R} = {\bf c}{\bf 1}_{L_h}^T, \, {\bf b}= d{\bf 1}_{n_b},{\bf c}\in\mathbb{R}^{L_m},d\in\mathbb{R}\},
\]
where we have assumed that the spectral response of the MS channels spans over the  entire $L_h$ HS spectral bands, and the HS bands are contiguous in frequency. The case in which the  spectral response of the MS channels spans over subsets of the $L_h$ HS spectral bands corresponds to a minor modification of the reasoning provided below. The case in which the HS bands are not contiguous is somewhat more elaborate, but would follow the same line of reasoning.

For any $({\bf R,b})\in {\bf A}$, we may write
\begin{equation}
\label{eq:obj_in_A}
\begin{split}
\sum_{j=1}^{n_h}  \Big\| \R {\bf Y}_{h,:j} - \mathbf{Y}_{m, j}\mathbf{b} \Big\|^2 
    & = \sum_{j=1}^{n_h} \Big\|y_{h,j}{\bf c}- {\bf y}_{m,j}d\Big\|^2,
\end{split}
\end{equation}
for some ${\bf c}\in\mathbb{R}^{L_m},d\in\mathbb{R}$ and 
where $y_{h,j} {\buildrel\rm def\over=} {\bf 1}_{L_h}^T{\bf Y}_{h,:j}$ and ${\bf y}_{m,j}{\buildrel\rm def\over=} \mathbf{Y}_{m, j}{\bf 1}_{n_b}$. Let us suppose that 
there exits a nonzero couple $({\bf c},d)$ nulling all the $n_h$ quadratic terms in  the right hand side of (\ref{eq:obj_in_A}). In this case, all vectors ${\bf y}_{m,j}$, for $j=1,\dots,n_h$ would be collinear with $\bf c$. Having into consideration that the 
components of   ${\bf y}_{m,j}$ represent the average intensities in the $L_m$ MS bands in the patch ${\bf P}_{c_j}$, such a scenario is highly unlikely, implying that the 
intersection of  the subspace $\bf A$  with the null space associated with the data term shown in the left hand side of (\ref{eq:obj_in_A}) is empty, except for the origin. We conclude, therefore,  that the Hessian of the quadratic objective function present in (\ref{eq:b_R_optim}) is positive definite and, thus, the solution  of the corresponding  optimization exists and  is unique.  An important  consequence  of
this uniqueness is that the subproblems (\ref{eq:est_B}) and  (\ref{eq:est_R}) have unique solutions; moreover, the system matrices present in the  expressions (\ref{eq:est_B_sol}) and (\ref{eq:est_R_sol}) are nonsingular.

\begin{figure}[tbh!]
\begin{center}
\colorbox{light}{\parbox{1.0\columnwidth}{
\begin{algorithmic}
\REQUIRE data: $\Y_h$ and $\Y_m$; regularization parameters: $\lambda_R$ and $\lambda_B$.
\STATE {Blur $\Y_m$ with a strong blur.}
\STATE {Blur $\Y_h$ with a correspondingly scaled blur.}
\STATE {Estimate $\R$ using~(\ref{eq:est_R_sol}) on the blurred data.}
\STATE {Estimate $\B$ using~(\ref{eq:est_B_sol}) on the original observed data.}
\STATE {Normalize $\mathbf{b}$ to unit DC gain.}
\end{algorithmic}
}}
\caption{{Summary of the method to estimate the spectral response $\R$ and the spatial blur; note that $\B$ and $\mathbf{b}$ are just two different ways of expressing this spatial blur. $\Y_m$ and $\Y_h$ refer to the multispectral and hyperspectral observations, respectively.}}
\label{alg:R_B}
\end{center}
\end{figure}

\section{Experimental study} \label{sec:exp}

In this section, we first describe the datasets that were used in the experimental tests, and the indices that were used to evaluate the quality of the results. We then give some details on the implementation of our algorithm and, finally, we present the experimental results, which include comparisons with several other fusion methods.

\subsection{Data sets}

Three datasets were used to test the different algorithms. Dataset A was purely synthetic. The ground truth image was a collection of simple geometric shapes composed of different hypothetical materials. In order to simulate the different materials, the U.S. Geological Survey Digital Spectral Library splib06 was used.\footnote{Available at \url{http://speclab.cr.usgs.gov/spectral-lib.html}.} This library assembles the reflectance values of different materials (e.g., minerals, plants, microorganisms, man-made materials) as measured by different instruments, covering the wavelength range from ultraviolet to far infrared. One of the instruments used to spectroscopically analyze the data was the National Aeronautics and Space Administration (NASA) Airborne Visible/Infra-Red Imaging Spectrometer (AVIRIS), which is capable of delivering calibrated images in 224 contiguous spectral channels within the 0.4-\SI{2.5}{\micro\metre} range~\cite{Kramer1994}.\footnote{More information is available at \url{http://aviris.jpl.nasa.gov/}.} Five signatures from this library were randomly selected as endmembers, and the image was built under the linear mixing model.

We created an image with high resolution both in the spatial and in the spectral domains, to serve as ground truth. To create a hyperspectral image, we spatially blurred the ground truth one, and then downsampled the result by a factor of 4 in each direction. Three different spatial blurs (block filter with dimensions $5 \times 5$, Gaussian filter with $\sigma = 2$ and support $5 \times 5$, and the Starck-Murtagh filter~\cite{Starck1994}) were used to synthesize three different HSIs. A false color representation of a hyperspectral image can be seen in Fig.~\ref{fig:Yhim_down_synth}, in which different colors correspond to different materials. To create panchromatic and multispectral images, the spectral response of the IKONOS satellite was used. This satellite captures both a panchromatic (0.45-\SI{0.90}{\micro\metre}) and four multispectral bands (0.45-0.52, 0.52-0.60, 0.63-0.69 and 0.76-\SI{0.90}{\micro\metre})~\cite{Kramer1994}. {Unless otherwise noted}, Gaussian noise was added to the hyperspectral image (SNR=30 dB) and to the multispectral image (SNR=40 dB).

Dataset B was semi-synthetic. It was based on a standard hyperspectral image (Pavia University, see Fig.~\ref{fig:pavia}). This image was obtained with the Reflective Optics System Imaging Spectrometer (ROSIS), which has 115 spectral bands, spanning the 0.43-\SI{0.86}{\micro\metre} spectral range, and a spatial resolution of \SI{1.3}{\metre}~\cite{Kramer1994}.\footnote{More information is available at~\url{http://messtec.dlr.de/en/technology/dlr-remote-sensing-technology-institute/hyperspectral-systems-airborne-rosis-hyspex/index.php}.} This image was used as ground truth. Hyperspectral, multispectral and panchromatic images were generated from it as described for dataset A. 

Dataset C consisted of images taken above Paris (see Fig.~\ref{fig:PAN_paris}), and was obtained by two instruments on board the Earth Observing-1 Mission (EO-1) satellite, the Hyperion instrument and the Advanced Land Imager (ALI). Hyperion is a hyperspectral imager with a spatial resolution of 30 meters; the ALI instrument provides both multispectral and panchromatic images at resolutions of 30 and 10 meters, respectively~\cite{Middleton2003}.\footnote{More information is available at \url{http://eo1.gsfc.nasa.gov/}, \url{http://eo1.usgs.gov/sensors/ali} and \url{http://eo1.usgs.gov/sensors/hyperioncoverage}.} The hyperspectral and panchromatic images were directly used for experiments on hyperspectral+panchromatic fusion, and therefore we had no access to the ground truth. For experiments on the fusion of hyperspectral and multispectral images, we needed the HSI to have lower resolution than the MSI, and therefore we first reduced the spatial resolution of the hyperspectral image by blurring with the Starck-Murtagh filter and downsampling, as described above for dataset A (using a downsampling factor of 3, in this case). The original hyperspectral image, before blurring and downsampling, was used as ground truth.

\subsection{Quality indices}

To evaluate the quality of fusion results, three indices taken from the literature were used, when a ground truth image was available, as was the case for datasets A and B, and for HS+MS fusion on dataset C. The first index was the \textit{Erreur Relative Globale Adimensionnelle de Synth\`{e}se} (ERGAS), proposed in~\cite{Wald2000} and defined, for an estimated image $\Z$ and a ground truth image $\widehat\Z$, as
\begin{equation}
\text{ERGAS}\big(\Z, \widehat\Z \big) \; {\buildrel\rm def\over=} \; 100 \frac{1}{S} \sqrt{\frac{1}{L_h} \sum_{l = 1}^{L_h}  \frac{\text{MSE}\big(\Z_{l:}, \widehat \Z_{l:}\big)}{\mu^2_{\widehat \Z_{l:}}}},
\end{equation}
where $S$ is the ratio between the resolutions of the hyperspectral image and of the multispectral or panchromatic one, i.e., $S = \sqrt{n_m/n_h}$; $\Z_{l:}$ and $\widehat \Z_{l:}$ are the $l$th bands of the estimated image and of the ground truth image, respectively; $\text{MSE}(\Z_{l:}, \widehat \Z_{l:})$ is the mean squared error between $\Z_{l:}$ and $\widehat \Z_{l:}$; and $\mu_{\widehat \Z_{l:}}$ is the mean of $\widehat \Z_{l:}$.

The second index was the Spectral Angle Mapper (SAM), which is the mean, among all pixels, of the angle between the vectors formed by the spectral representation of the pixel in the estimated image and the spectral representation of the same pixel in the ground truth image,
\begin{equation}
\text{SAM}\big(\Z, \widehat\Z \big) \; {\buildrel\rm def\over=} \; \frac 1 {n_m} \sum_{j=1}^{n_m} \arccos \left(\frac{\Z_{: j}^T \, \widehat\Z_{: j}}{\big\|\Z_{: j}\big\|_2 \, \big\|\widehat\Z_{: j}\big\|_2}\right),
\end{equation}
where $\Z_{: j}$ denotes the spectral representation of the $j$th pixel of the estimated image and $\widehat\Z_{: j}$ denotes the same for the ground truth image. This index is an indicator of the spectral quality of the estimated image. In this paper we report the value of the SAM index in degrees.

The third index was based on the Universal Image Quality Index (UIQI), proposed by Wang \textit{et al.}~\cite{Wang2002}. It was computed on a sliding window of size $32 \times 32$ pixels, and averaged over all window positions. Denoting by $\mathbf z_i$ the $i$th windowed segment of a single-band image and by $\widehat{\mathbf z}_i$ the corresponding segment of a single-band ground truth image, the UIQI is given by
\begin{equation} \label{eq:uiqisingleband}
\text{Q}(\mathbf z, {\widehat{\mathbf z}}) \; {\buildrel\rm def\over=} \; \frac 1 M \sum_{i=1}^M \frac{\sigma_{\mathbf z_i \widehat{\mathbf z}_i}} {\sigma_{{\mathbf z}_i} \sigma_{\widehat{\mathbf z}_i}} \times \frac{2 \, \mu_{{\mathbf z}_i} \mu_{\widehat{\mathbf z}_i}}{\mu_{{\mathbf z}_i}^2 + \mu_{\widehat{\mathbf z}_i}^2} \times \frac{2 \, \sigma_{{\mathbf z}_i} \sigma_{\widehat{\mathbf z}_i}}{\sigma_{{\mathbf z}_i}^2 + \sigma_{\widehat{\mathbf z}_i}^2},
\end{equation}
where $M$ is the number of window positions, $\sigma_{\mathbf z_i \widehat{\mathbf z}_i}$ is the covariance between $\mathbf z_i$ and $\widehat{\mathbf z}_i$, $\sigma_{{\mathbf z}_i}$ is the standard deviation of $\mathbf z_i$, and $\sigma_{\widehat{\mathbf z}_i}$ is the standard deviation of $\widehat{\mathbf z}_i$. This index has a range of $[-1,1]$, being equal to 1 when $\mathbf z = {\widehat{\mathbf z}}$.

The definition of the UIQI index was extended to multiband hyperspectral images by simple averaging:
\begin{equation} \label{eq:uiqi}
\text{UIQI}\big(\Z, {\widehat\Z}\big) \; {\buildrel\rm def\over=} \; \frac 1 {L_h} \sum_{l=1}^{L_h} \text{Q}\big(\Z_{l:}, \widehat \Z_{l:}\big).
\end{equation}
$\text{Q}$ was computed using the MATLAB code provided by Wang \textit{et al.}\footnote{Available from \url{https://ece.uwaterloo.ca/~z70wang/research/quality_index/demo.html}.}

When working with the fusion of hyperspectral and panchromatic images from dataset C, we had no access to the ground truth. We hence only show false color representations of the estimated images, for visual inspection.

\subsection{Implementation details} \label{sec:implementation_details}

In the experimental tests, we performed two preprocessing steps on the hyperspectral data: First, uncalibrated or very noisy bands were removed{: when information on which bands were uncalibrated was available (as in the case of Hyperion), it was used; very noisy bands were identified manually.} Second, the data were denoised by projecting $\Y_h$ onto a subspace of dimension $L_s=10$ found through truncated SVD; the ground truth images, when available, were also projected onto this subspace. Making $L_s=10$ allowed us preserve at least $99.95\%$ of the energy of the original images from all datasets. Dataset C consisted of raw data, in which the energy per band strongly varied across the spectrum. For this reason, before denoising, we normalized all bands of this dataset so that the 0.999 intensity quantile corresponded to a value of 1. {A summary of these two steps can be seen} in Fig.~\ref{alg:preprocessing}. 

\begin{figure}[tbh!]
\begin{center}
\colorbox{light}{\parbox{1.0\columnwidth}{
\begin{algorithmic}
\REQUIRE observed hyperspectral and multispectral data.
\STATE {Remove uncalibrated/noisy bands.}
\STATE {Normalize of each band.}
\STATE {Denoise the data.}
\RETURN {$\Y_h$ and $\Y_m$ .}
\end{algorithmic}
}}
\caption{{Summary of the preprocessing steps.}}
\label{alg:preprocessing}
\end{center}
\end{figure}

\begin{figure*}[!t]
	\centering
	\subfloat[ERGAS.]{\includegraphics[scale=.33]{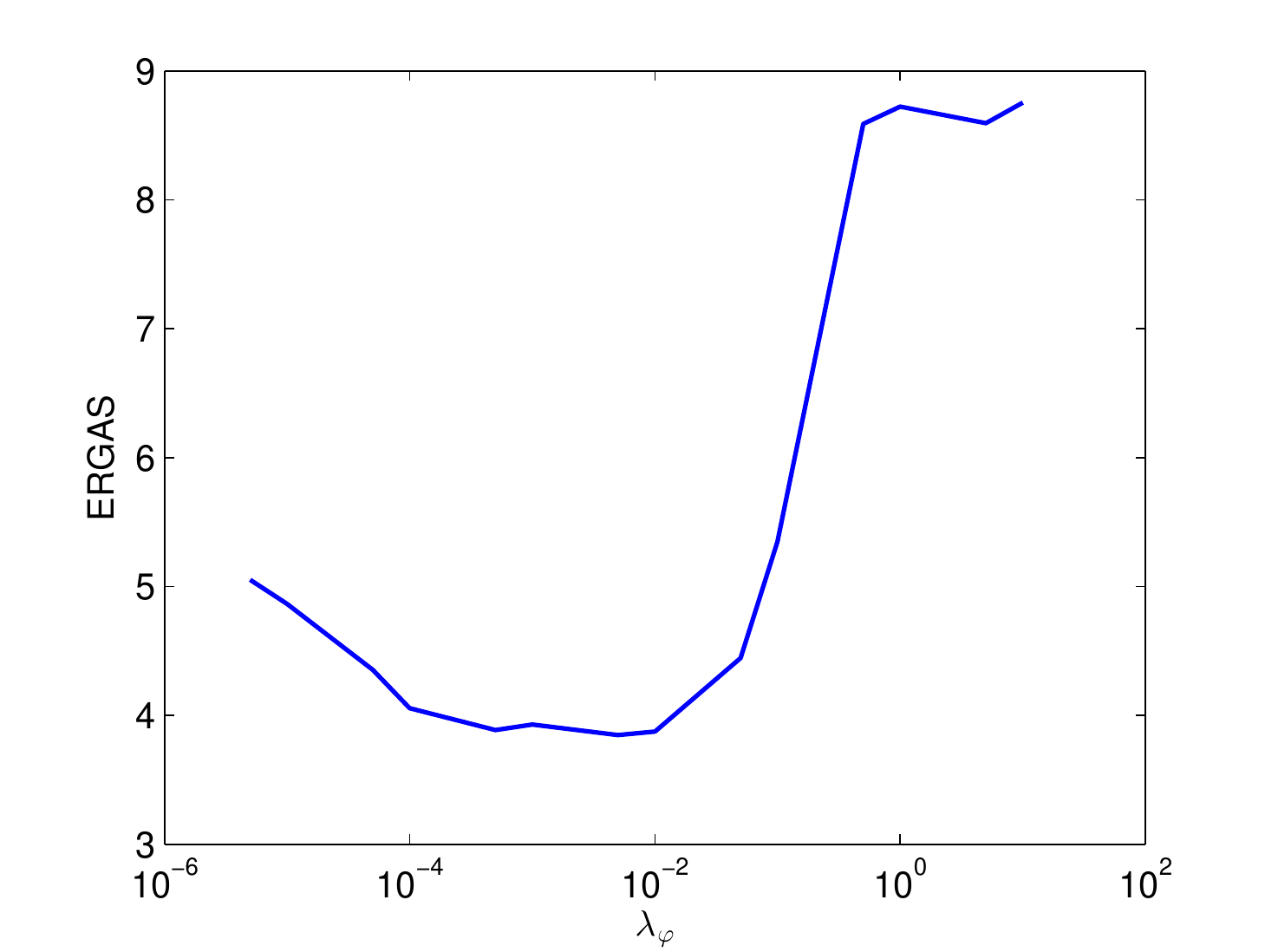}%
		\label{fig:lambda_varphi_ergas}}
	\hfil
	\subfloat[SAM.]{\includegraphics[scale=.33]{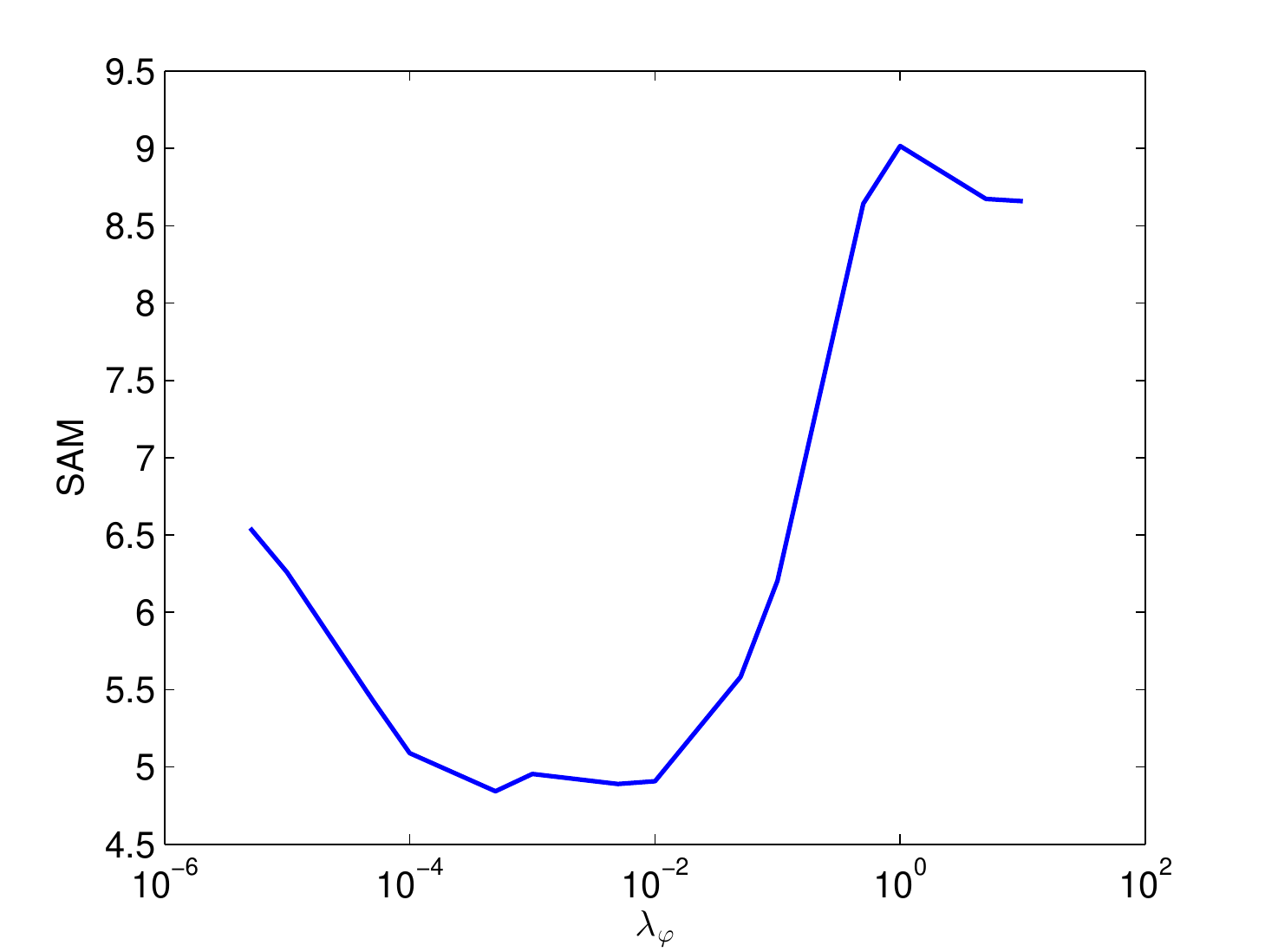}%
		\label{fig:lambda_varphi_sam}}
	\hfil
	\subfloat[UIQI.]{\includegraphics[scale=.33]{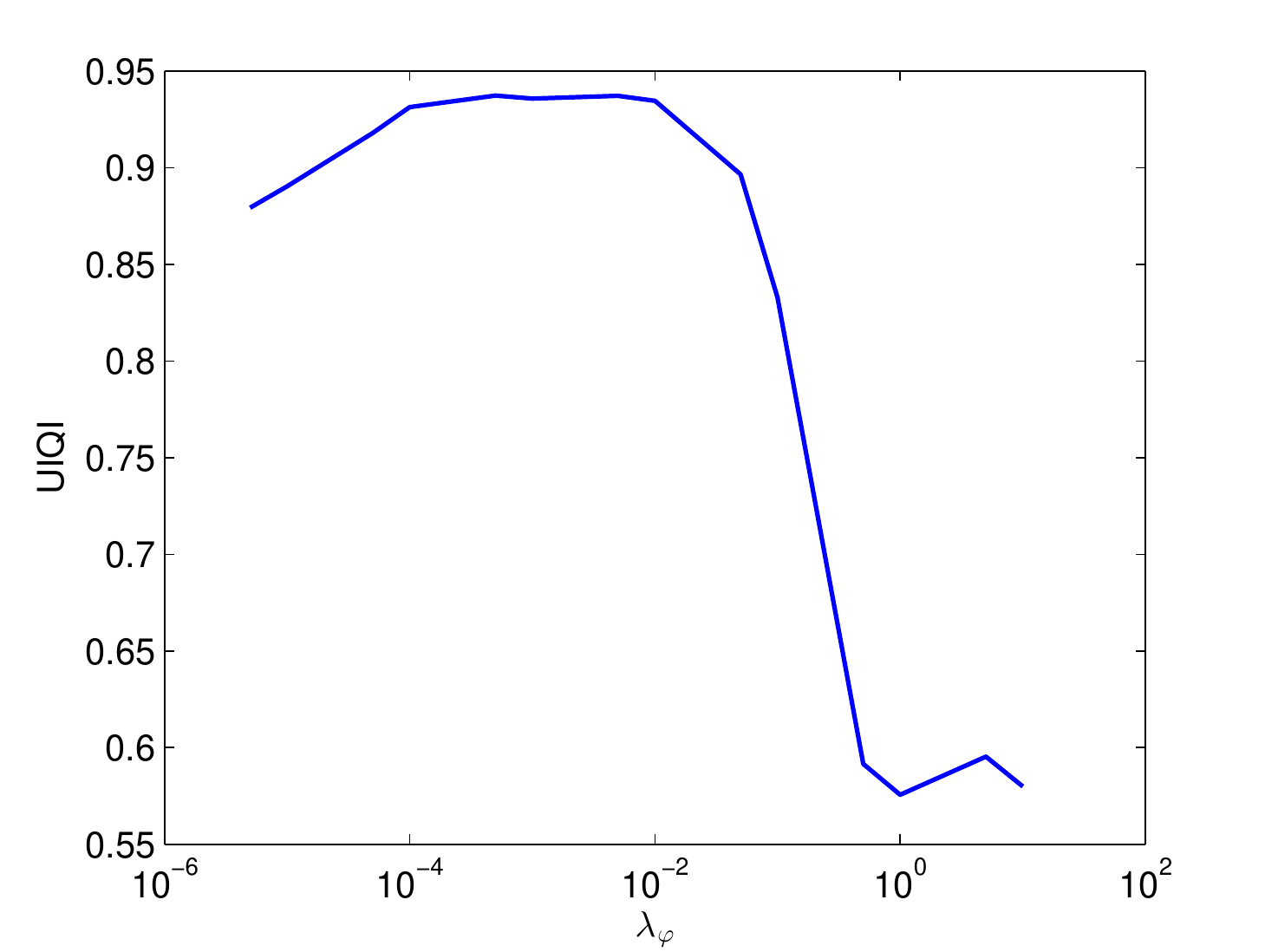}%
		\label{fig:lambda_varphi_uiqi}}
	\caption{{Quality indices for different values of $\lambda_{\varphi}$, for dataset B (HSI+PAN) fusion.}}
	\label{fig:lambda_varphi}
\end{figure*}

\begin{figure*}[!t]
	\centering
	\subfloat[Spatial blur between the HSI and the panchromatic band.]{\includegraphics[scale=.30]{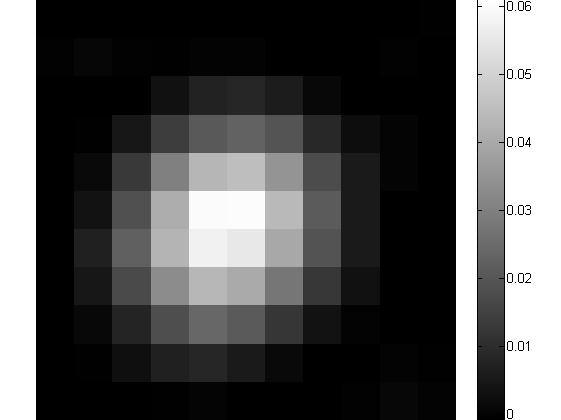}%
		\label{fig:B_hs_pan}}
	\hfil
	\subfloat[Spectral relationship between the HSI and the panchromatic band.]{\includegraphics[scale=.36]{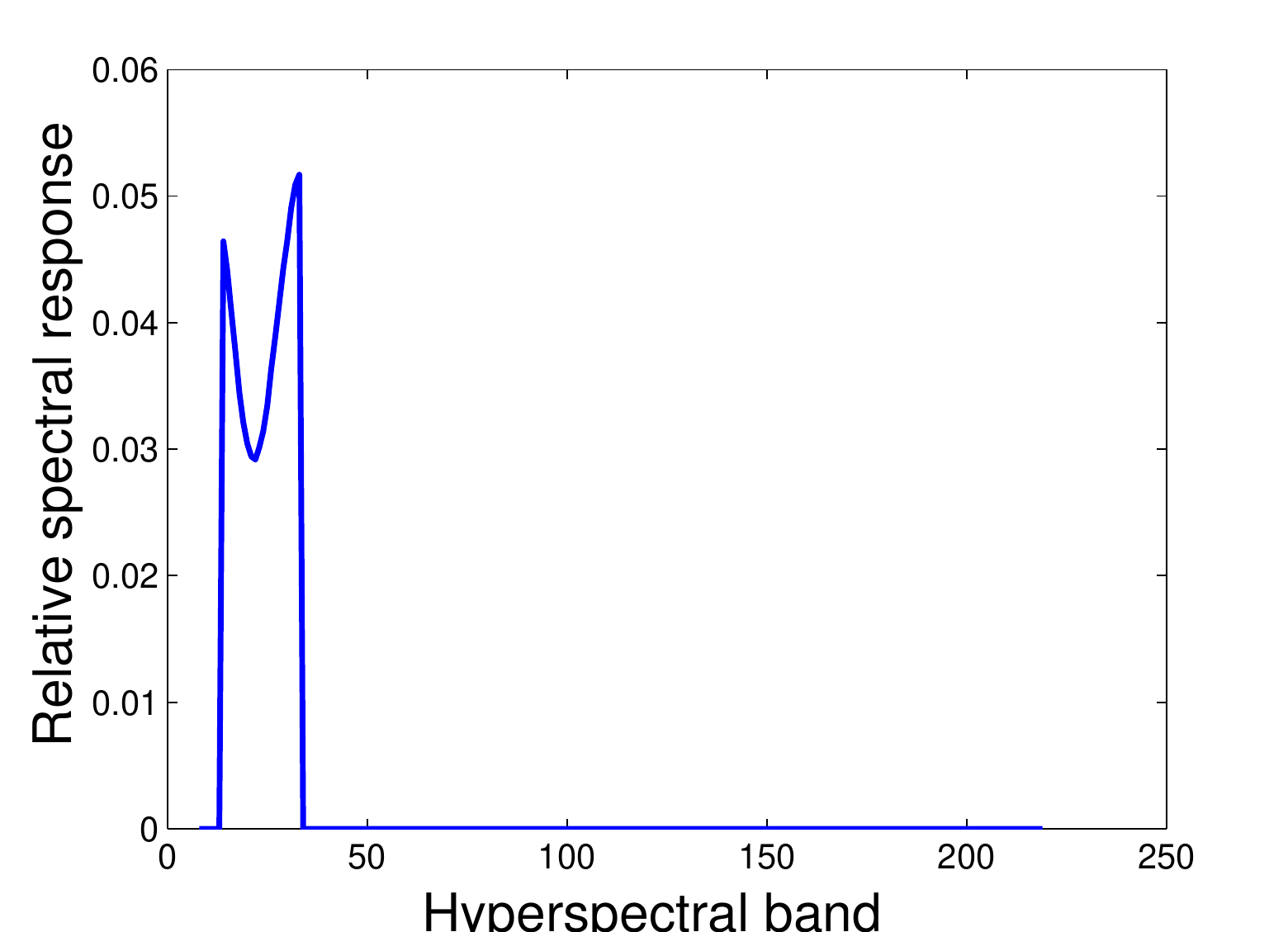}%
		\label{fig:R_hs_pan}}
	\hfil
	\subfloat[Spectral relationship between the HSI and the MSI. Different multispectral bands are shown in different colors.]{\includegraphics[scale=.36]{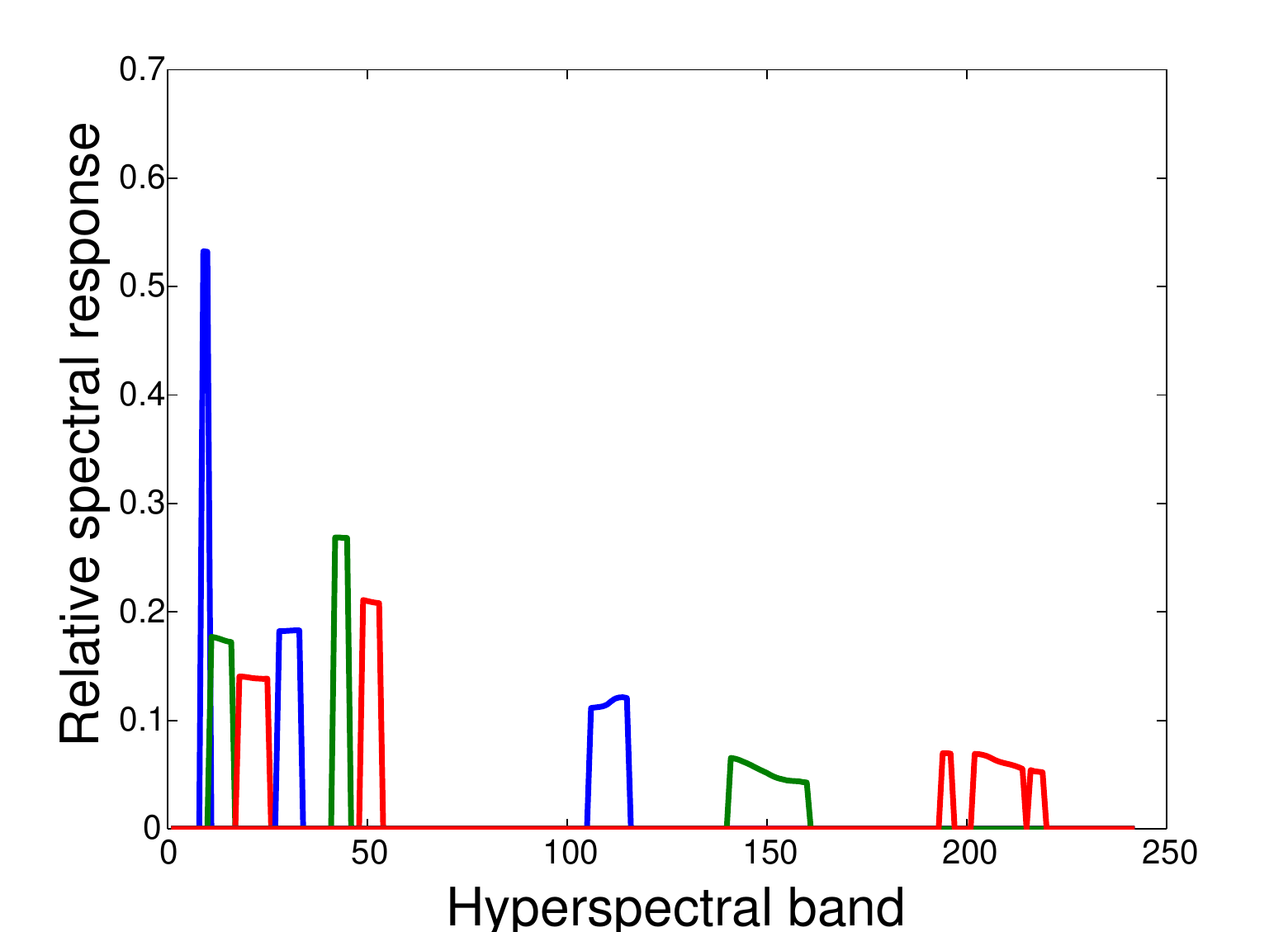}%
		\label{fig:R_hs_ms}}
	\caption{{Spectral and spatial blur estimates for dataset C.}}
	\label{fig:parisRB}
\end{figure*}

After the preprocessing, we estimated the spectral and spatial responses as described in Section~\ref{sec:estimate}. We then estimated matrix $\E$ using VCA;\footnote{Available from \url{http://www.lx.it.pt/~bioucas/code/demo_vca.zip}.} since the subspace estimated by VCA shares the dimension of the subspace estimated by SVD~\cite{Nascimento2005}, we also made $L_s=10$ in this step. Since VCA has a random component, we performed ten runs of our algorithm in each case, and we report the average of the corresponding results. Their standard deviation was negligible.

{The tuning of the values of the algorithm's parameters is an interesting and complex topic, with a number of techniques that can be adapted to problems such as this. Two examples of these techniques are Stein's Unbiased Risk Estimator (SURE)}~\cite{Donoho1995} {and Generalized Cross-Validation (GCV)}~\cite{Golub1979}. {We verified experimentally, however, that as long as our remote sensing images were preprocessed as described earlier, constant values for these parameters tended to lead to near-optimal results. To choose these values,} we first found the optimal values for each situation, and computed the corresponding quality indices. We then chose a set of parameter values that were the same for all situations, but that yielded quality indices that were very close to the previously found optimal ones. These values were $\lambda_m = 1$ and $\mu=5 \times 10^{-2}$. We used $\lambda_{\varphi}=10^{-2}$ when fusing a HSI with a panchromatic image and $\lambda_{\varphi}=5 \times 10^{-4}$ when fusing a HSI with a MSI. We used $\lambda_B = \lambda_R = 10$ (see Section~\ref{sec:exp_R:_B} for how these values were chosen). {To illustrate the influence of different values of $\lambda_{\varphi}$ on the ERGAS, SAM and UIQI quality indices (including the situation when there is no regularization, i.e., $\lambda_{\varphi}=0$), we performed a series of experiments on the HSI+PAN fusion of dataset B (see Section~{\ref{sec:exp_B}} for more details). The results of these experiments can be seen in Fig.~{\ref{fig:lambda_varphi}}; for $\lambda_{\varphi}=0$, the values of ERGAS, SAM and UIQI were 5.046, 6.587 and 0.882, respectively}.

\begin{figure*}[!t]
	\centering
	\subfloat[Panchromatic image.]{\includegraphics[scale=.333, trim=0 1 0 0, clip=true]{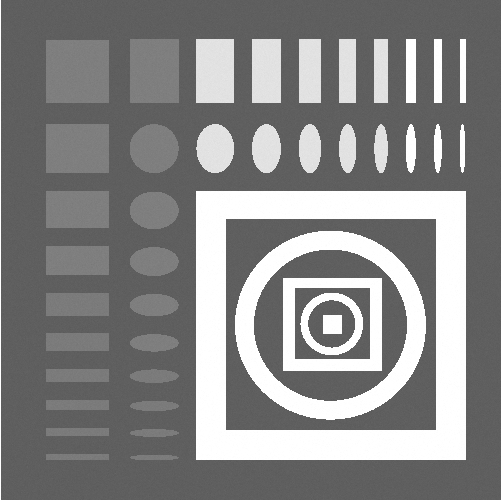}%
		\label{fig:Ymim_synth}}
	\hfil
	\subfloat[Hyperspectral image (false color).]{\includegraphics[scale=1]{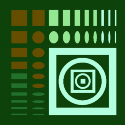}%
		\label{fig:Yhim_down_synth}}
	\hfil
	\subfloat[HySure's result (false color).]{\includegraphics[scale=.25]{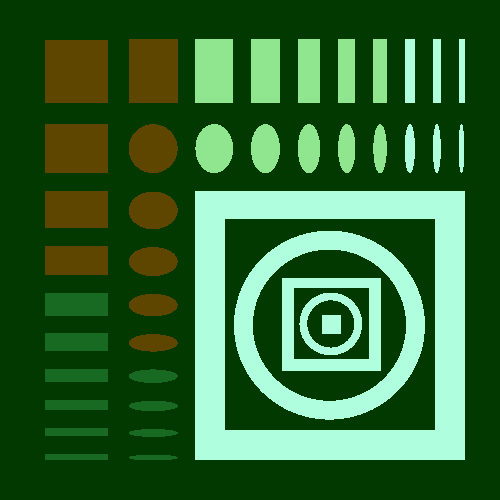}%
		\label{fig:Zim_synth}}
	\caption{Hyperspectral + panchromatic fusion on dataset A.}
	\label{fig:synth}
\end{figure*}

In~\cite{Boyd2011}, a stopping criterion was proposed for problems solved via ADMM. We verified that this criterion worked well, always yielding less than 200 iterations. Given this, we ran the algorithm for 200 iterations in every case.

\subsection{Experimental results}

\subsubsection{Estimation of the spatial and spectral responses of the sensors} \label{sec:exp_R:_B}

Our first experiments were aimed at testing the estimation of the spectral and spatial responses of the sensors on real-life data. After checking that the results on datasets A and B were rather accurate, we chose the values of $\lambda_B$ and $\lambda_R$ that yielded the highest-quality results on those datasets. We then tested the estimation method, with those parameter values, on dataset C. In the estimation of the spectral response, we took into account the available information on the overlap between the hyperspectral bands and the multispectral and panchromatic ones. Since the original hyperspectral and multispectral images of this dataset have the same resolution, for that pair of images we have set $\B=\mathbf{I}$, corresponding to no spatial blur, and we have just estimated $\R$, without applying any additional spatial blur. For the hyperspectral+panchromatic images, which have different resolutions, we performed the estimation as described in Fig.~\ref{alg:R_B}. The estimated blurs, which look quite reasonable, can be seen in Fig.~\ref{fig:parisRB}.

\subsubsection{Fusion of hyperspectral and panchromatic images}  \label{sec:exp_B}

A number of methods for the fusion of multiband images with panchromatic ones, drawn from the pansharpening literature, were used for comparison with HySure. Those methods were originally built having in mind the fusion of panchromatic images with multispectral ones, i.e., they were built for a small number of bands, and not for the large number of bands of a typical hyperspectral image. The methods can, however, be extended in a straightforward manner to hyperspectral images, since they have no restrictions on the number of bands. In what follows, a quick rundown of those methods is given. A criterion used to choose the methods for comparison was that they should not impose restrictions on the ratio between the resolutions of the high spatial resolution image and the low spatial resolution one. Since our method does not impose such restrictions, we only compared it against similarly built methods.

In~\cite{Amro2011}, Amro \textit{et al.} divided the pansharpening methods into several categories. One of them is the Component Substitution family; different methods from this family were tested in this work. They are characterized by the transformation of the multispectral bands into a set of components, usually through a linear transformation. After this, a component of the transformed multiband image is replaced with an image derived from the panchromatic one, and then the transformation is undone. These methods work well only when the spectra of the two data sources almost overlap, a condition which may not be fulfilled when fusing panchromatic and hyperspectral images. The Gram-Schmidt adaptive (GSA) method from Aiazzi \textit{et al.}~\cite{Aiazzi2007} is an adaptation of the Gram-Schmidt spectral sharpening method (GS). The latter is based on the Gram-Schmidt transformation of the different low spatial resolution bands, followed by the substitution of the first band of the transformed image with a modified version of the panchromatic band. This modified version is given by a weighted sum of the multispectral bands, expanded to the spatial resolution of the panchromatic image. The weights are obtained in different ways, and that is the main difference between GS and GSA. In GS, they are assumed to be the same for all bands, while in GSA they are estimated from the observed data, usually guaranteeing better results. In the case of fusion with hyperspectral images, GSA involves the inversion of a matrix that is close to singular, possibly affecting the quality of the results. Nevertheless, as will be seen later, the experiments showed an improvement of GSA relative to GS. The Fast Intensity-Hue-Saturation Fusion Technique (FIHS) is another method included in this family. It is similar to GSA and GS, with the difference that the processing is made in the IHS color space, with the panchromatic image replacing the intensity component of the multiband image~\cite{Tu2004}. {Another method relies on the Principal Component Analysis (PCA) of the multiband image}, and replaces the first principal component with the panchromatic image~\cite{Tu2004}.

Another family of methods is the Relative Spectral Contribution family. An example is the Brovey Transform method (BT), based on the chromaticity transform~\cite{Tu2004}. In this method, each pixel of the estimated image is given by the corresponding pixel of the panchromatic image, weighted by a linear combination of the values of the different spatially expanded multispectral bands for this same pixel. Finally, another family is the High-Frequency Injection one, from which we used the Box High-pass Filtering method (HPF). It is characterized by the extraction of high frequency information from the high spatial resolution image, followed by the injection of this information into the multiband image~\cite{Amro2011}. To perform the high frequency extraction, the method starts by producing a low-pass version of the panchromatic image through a box filtering operation. This blurred image is then subtracted from the original one, yielding a high frequency version of it.

\begin{figure*}[!t]
	\centering
	\subfloat[Observed HSI (false color).]{\includegraphics[scale=1.2]{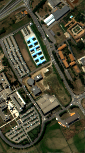}%
		\label{fig:paviahs}}
	\hfil
	\subfloat[Observed panchromatic image.]{\includegraphics[scale=.4]{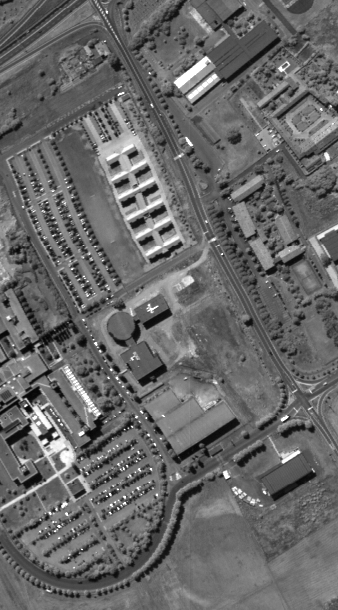}%
		\label{fig:paviapan}}
	\hfil
	\subfloat[HySure's result (false color).]{\includegraphics[scale=.3]{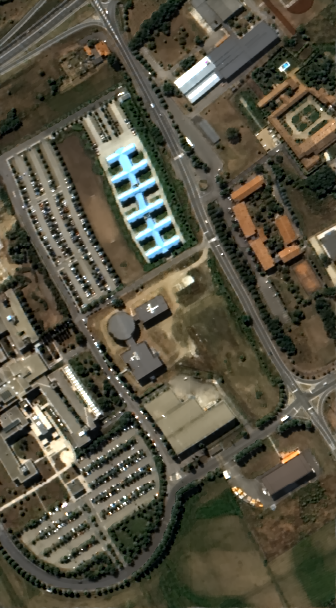}%
		\label{fig:Zimhat_Pavia_pan}}
	\hfil
	\subfloat[BT's result (false color).]{\includegraphics[scale=.3]{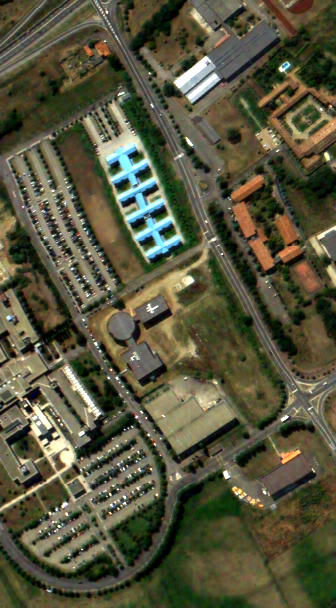}%
		\label{fig:Zimhat_Brovey_Pavia_pan}}
	\hfil
	\subfloat[Evolution of the cost function during the optimization.]{\includegraphics[scale=.48]{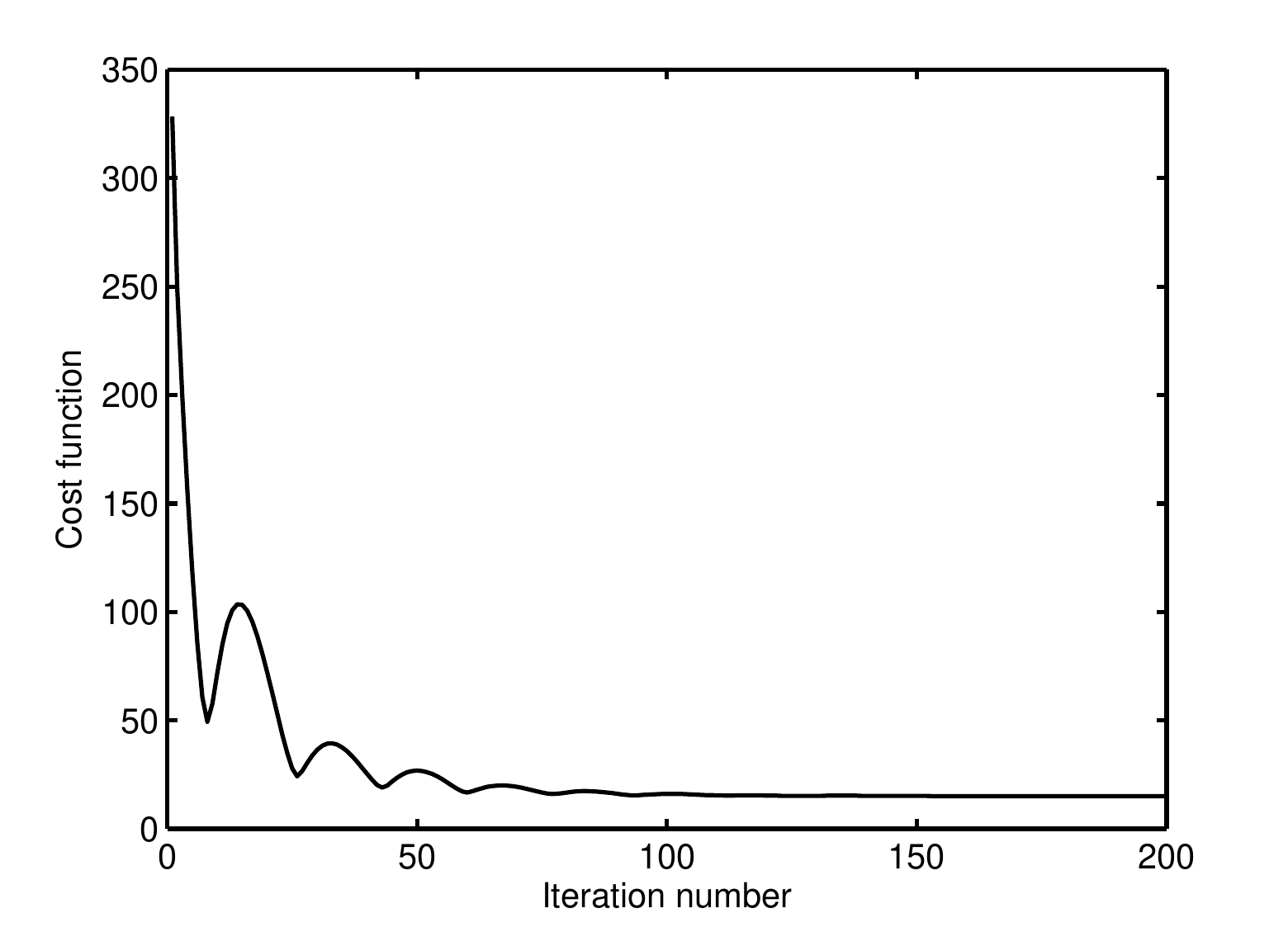}%
		\label{fig:costf}}
	\hfil
	\subfloat[Root-mean-square error (RMSE) between the estimated image and the ground truth, for the different bands (for the three best methods).]{\includegraphics[scale=.45]{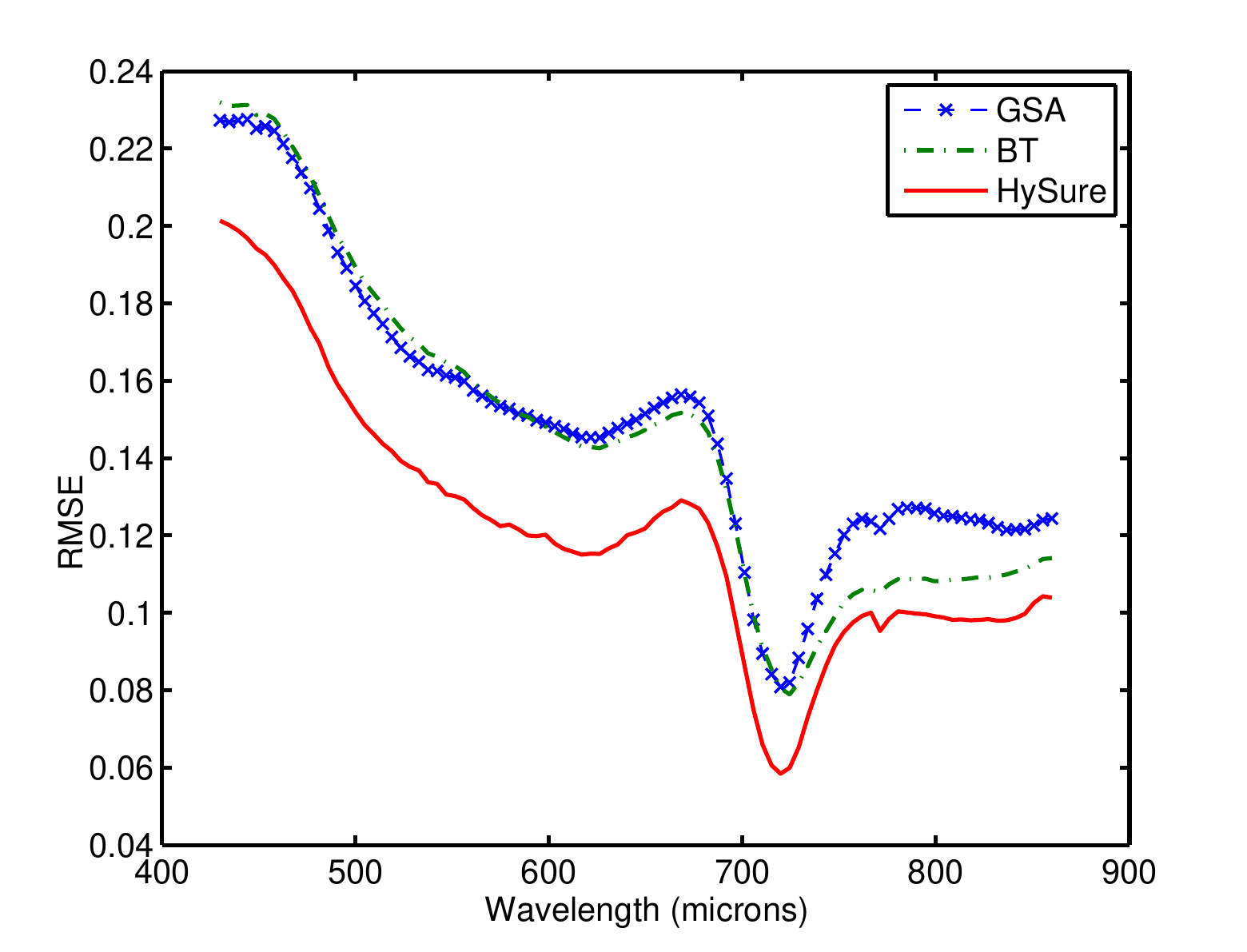}%
		\label{fig:spectrum}}
	\hfil
	\caption{Results for dataset B (HSI+PAN fusion).}
	\label{fig:pavia}
\end{figure*}

\begin{figure*}[!t]
\centering
\subfloat[Observed panchromatic image.]{\includegraphics[scale=.666, trim=16 16 0 0, clip=true]{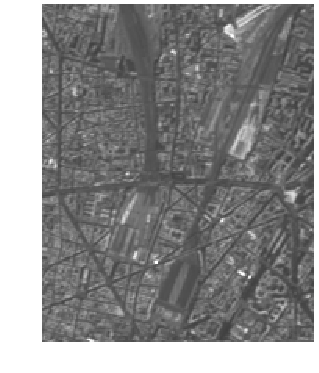}%
\label{fig:PAN_paris}}
\hfil
\subfloat[{Observed HSI.}]{\includegraphics[scale=1.5]{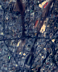}%
\label{fig:Yhim_paris}}
\hfil
\subfloat[HySure's result.]{\includegraphics[scale=.5]{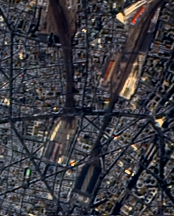}%
\label{fig:Zimhat}}
\hfil
\subfloat[GSA's result.]{\includegraphics[scale=.5]{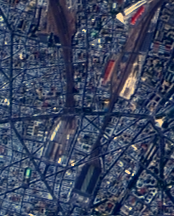}%
\label{fig:Zimhat_GSA}}
\hfil
\subfloat[GS's result.]{\includegraphics[scale=.5]{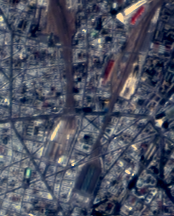}%
\label{fig:Zimhat_GS1}}
\hfil
\subfloat[HPF's result.]{\includegraphics[scale=.5]{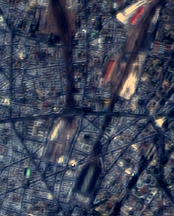}%
\label{fig:Zimhat_Box}}
\hfil
\subfloat[BT's result.]{\includegraphics[scale=.5]{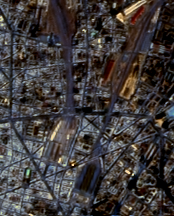}%
\label{fig:Zimhat_Brovey}}
\hfil
\subfloat[FIHS's result.]{\includegraphics[scale=.5]{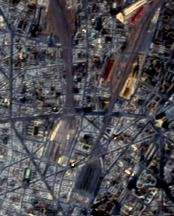}%
\label{fig:Zimhat_FIHS}}
\hfil
\subfloat[PCA's result.]{\includegraphics[scale=.5]{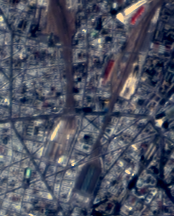}%
\label{fig:Zimhat_PCA}}
\caption{Results for dataset C (HSI+PAN fusion). All images, except (a), are in false color.}
\label{fig:paris}
\end{figure*}

Fig.~\ref{fig:synth} and Table~\ref{tab:synth} show the results of the various methods for dataset A. We only show the results for the Starck-Murtagh blur, since the results for the other two blurs were very similar to these. The results for dataset B (again, just for the Starck-Murtagh blur) can be seen in Fig.~\ref{fig:pavia} and Table~\ref{tab:pavia_hs_pan}. The evolution of the cost function~(\ref{eq:optimizationproblem}) during the optimization is shown in Fig.~\ref{fig:costf}. Fig.~\ref{fig:spectrum} shows {the root-mean-square error (RMSE) between the estimated image and the ground truth as a function of band wavelength} for the three best methods. Dataset C allowed us to evaluate the methods on real-life data. Fig.~\ref{fig:paris} shows the results. 

\begin{table}[!t]
\renewcommand{\arraystretch}{1.3}
\caption{Results for dataset A (HSI+PAN fusion).}
\label{tab:synth}
\centering
\begin{tabular}{l||c|c|c}
\multicolumn{1}{r}{} & ERGAS & SAM & UIQI\\
\hline
\hline
GS & 1.330 & 1.136 & 0.868 \\ 
 \hline 
GSA & 1.268 & 1.156 & 0.870 \\ 
 \hline 
FIHS & 1.788 & 1.456 & 0.863 \\ 
 \hline 
PCA & 1.451 & 1.149 & 0.865 \\ 
 \hline 
BT & 1.832 & 1.427 & 0.875 \\ 
 \hline 
HPF & 3.277 & 1.688 & 0.845 \\ 
 \hline 
 \textbf{HySure} & \textbf{0.717} & \textbf{0.524} & \textbf{0.895}\\
 \hline
\hline
\end{tabular}
\end{table}

\begin{table*}
	\renewcommand{\arraystretch}{1.3}
	\caption{Results for dataset B (HSI+PAN fusion).}
	\label{tab:pavia_hs_pan}
	\centering
	\begin{tabular}{l||c|c|c||c|c|c||c|c|c}
		SNR($\Y_m$) & \multicolumn{3}{c||}{40 dB} & \multicolumn{3}{c||}{30 dB} & \multicolumn{3}{c}{20 dB}\\
		\hline
		SNR($\Y_h$) & \multicolumn{3}{c||}{30 dB} & \multicolumn{3}{c||}{20 dB} & \multicolumn{3}{c}{20 dB}\\
		\cline{2-10}
		\multicolumn{1}{r||}{} & ERGAS & SAM & UIQI & ERGAS & SAM & UIQI & ERGAS & SAM & UIQI\\
		\hline
		\hline
		GS & 4.960 & 5.494 & 0.897 & 5.112 & 5.914 & 0.885 & 5.879 & 6.303 & 0.838\\ 
		\hline 
		GSA & 4.587 & 5.116 & 0.905 & 4.733 & 5.538 & 0.893 & 5.448 & 5.804 & 0.848\\ 
		\hline 
		FIHS & 4.813 & 5.255 & 0.905 & 4.962 & 5.669 & 0.894 & 5.661 & 5.908 & 0.848\\ 
		\hline 
		PCA & 7.609 & 9.448 & 0.774 & 7.712 & 9.711 & 0.766 & 8.280 & 10.020 & 0.730\\ 
		\hline 
		BT & 4.533 & \textbf{4.550} & 0.926 & 4.684 & 4.989 & 0.915 & 5.311 & \textbf{4.983} & 0.874\\ 
		\hline 
		HPF & 5.573 & 6.151 & 0.880 & 5.731 & 6.631 & 0.867 & 6.527 & 7.111 & 0.814\\ 
		\hline
		\textbf{HySure} & \textbf{3.813} & 4.856 & \textbf{0.937} & \textbf{3.894} & \textbf{4.938} & \textbf{0.933} & \textbf{4.630} & 5.507 & \textbf{0.897}\\
		\hline
		\hline
	\end{tabular}
\end{table*}

The proposed method outperformed the other ones in all cases, except for the SAM index in dataset B, in which it was surpassed by BT. We found that most published pansharpening methods seem to not deal well with the fact that the panchromatic image's spectral range does not overlap a large number of hyperspectral channels.

\begin{figure*}[!t]
	\centering
	\subfloat[Observed multispectral image.]{\includegraphics[scale=.5]{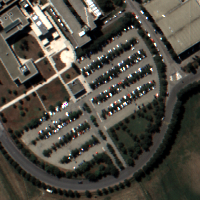}%
		\label{fig:Ymim_pavia_ms_hs}}
	\hfil
	\subfloat[Observed hyperspectral image.]{\includegraphics[scale=2]{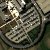}%
		\label{fig:Yhim_pavia_ms_hs}}
	\hfil
	\subfloat[HySure's result.]{\includegraphics[scale=.5]{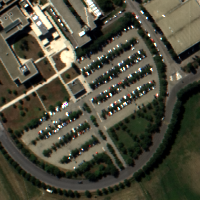}%
		\label{fig:Zimhat_pavia_ms}}
	\hfil
	\subfloat[ZBS's result.]{\includegraphics[scale=.5]{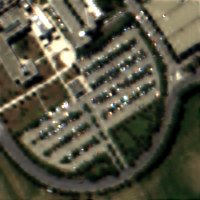}%
		\label{fig:Zimhat_Zhang_pavia_ms}}
	\caption{Results for dataset B (HSI+MSI fusion). All images are in false color. Figs.~\ref{fig:Zimhat_pavia_ms} and~\ref{fig:Zimhat_Zhang_pavia_ms} are very similar to Fig.~\ref{fig:Ymim_pavia_ms_hs} due to the false color rendering, but they have 93 bands, while Fig.~\ref{fig:Ymim_pavia_ms_hs} has only four.}
	\label{fig:pavia_ms}
\end{figure*}

\subsubsection{Fusion of hyper- and multispectral images}   \label{sec:exp_C}

The literature on the fusion of hyperspectral and multispectral images is much sparser than the one on pansharpening. As a consequence, we were only able to perform comparisons with one published method: we had access to an implementation of a method by Zhang \textit{et al.}~\cite{Zhang2009} (henceforth, designated by ZBS) and used it for comparisons on datasets B and C. This implementation needed the input HSIs and MSIs to be represented with the same spatial resolution. Therefore, we upsampled the HSIs to the resolution of the MSIs, using bicubic interpolation, for input to ZBS. This method does not estimate the spatial blur, needing it to be specified; we estimated it as in our method, with the difference that we worked with the upsampled version of the HSI. Following the lead of that method's authors, we chose the decomposition level of the Nondecimated Wavelet Transform to be three.

\begin{figure*}[!t]
\centering
\subfloat[{Observed multispectral image.}]{\includegraphics[scale=1]{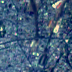}%
\label{fig:paris_ms_hs_ym}}
\hfil
\subfloat[{Observed hyperspectral image.}]{\includegraphics[scale=3]{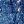}%
\label{fig:paris_ms_hs_yh}}
\hfil
\subfloat[{HySure's result.}]{\includegraphics[scale=1]{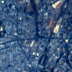}%
\label{fig:paris_ms_hs_nosso}}
\hfil
\subfloat[{ZBS's result.}]{\includegraphics[scale=1]{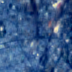}%
\label{fig:paris_ms_hs_zhang}}
\caption{{Results for dataset C (HSI+MSI fusion). All images are in false color. Figs.~{\ref{fig:paris_ms_hs_nosso}} and~{\ref{fig:paris_ms_hs_zhang}} are very similar to Fig.~{\ref{fig:paris_ms_hs_ym}} due to the false color rendering, but they have 128 bands, while Fig.~{\ref{fig:paris_ms_hs_ym}} has only nine.}}
\label{fig:paris_ms_hs}
\end{figure*}

\begin{figure*}[!t]
	\centering
	\begin{minipage}{0.49\linewidth}
		\centering
	\subfloat[{RMSE between the the results of both methods and the ground truth, per pixel, for dataset B.}]{\includegraphics[scale=.4]{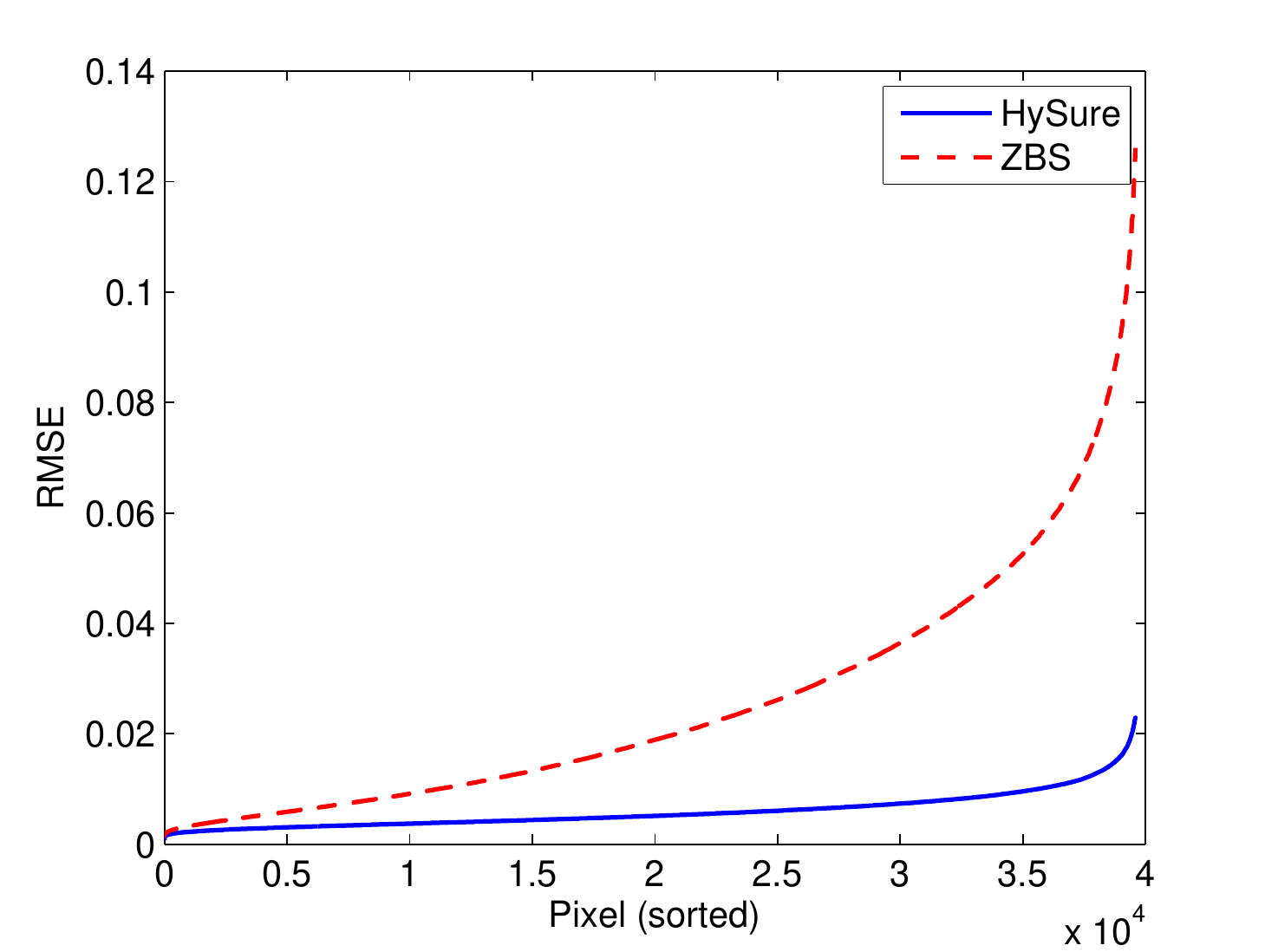}%
	\label{fig:rmse_plot_hs_ms_sorted_pavia}}
\end{minipage}
	\begin{minipage}{0.49\linewidth}
		\centering
	\subfloat[{RMSE between the the results of both methods and the ground truth, per pixel, for dataset C.}]{\includegraphics[scale=.4]{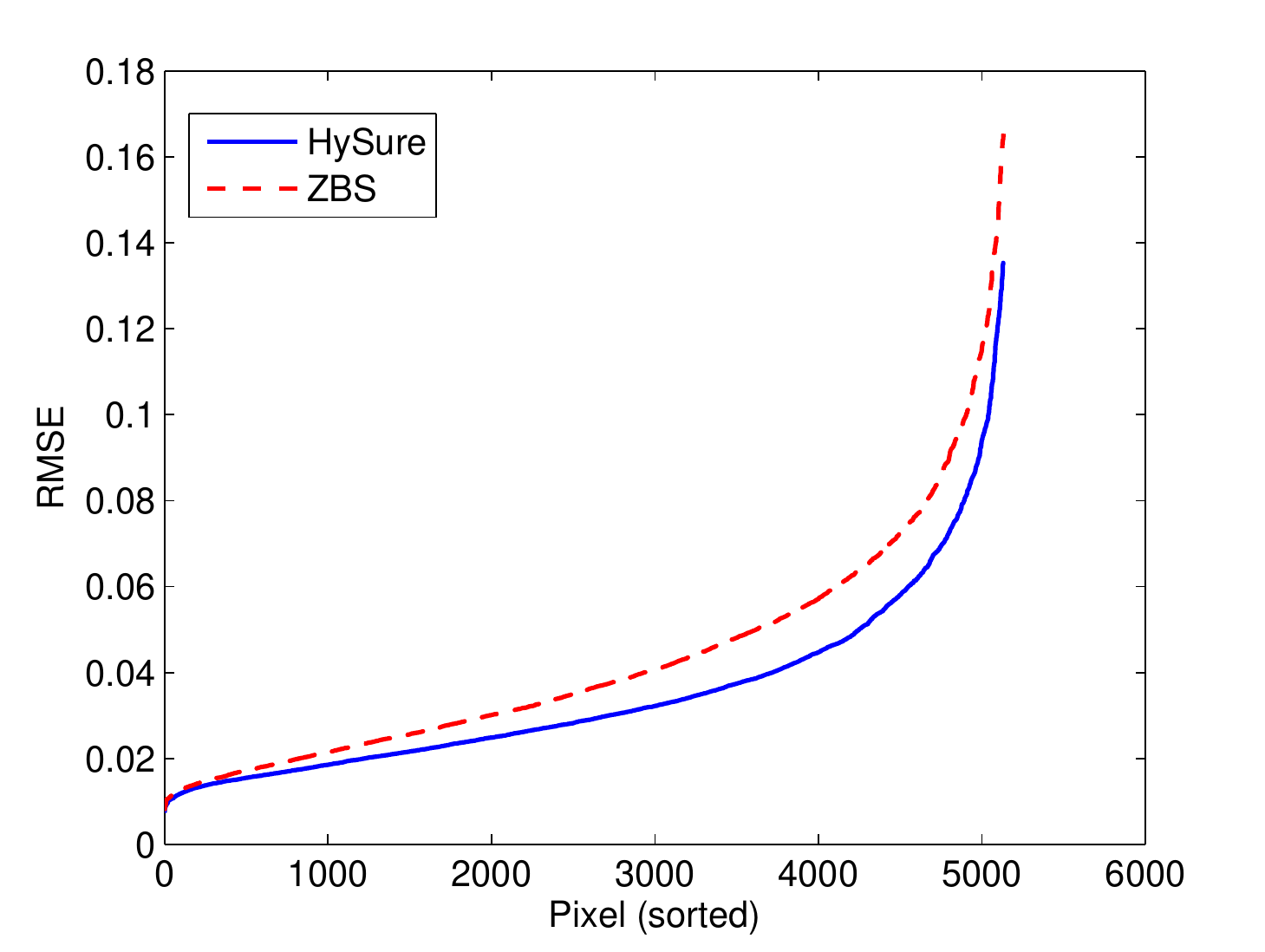}%
	\label{fig:rmse_plot_hs_ms_sorted}}
\end{minipage}
	\begin{minipage}{0.33\linewidth}
				\centering
	\subfloat[{Reflectance of the pixel corresponding to the 10th percentile of the RMSE (dataset B).}]{\includegraphics[scale=.33]{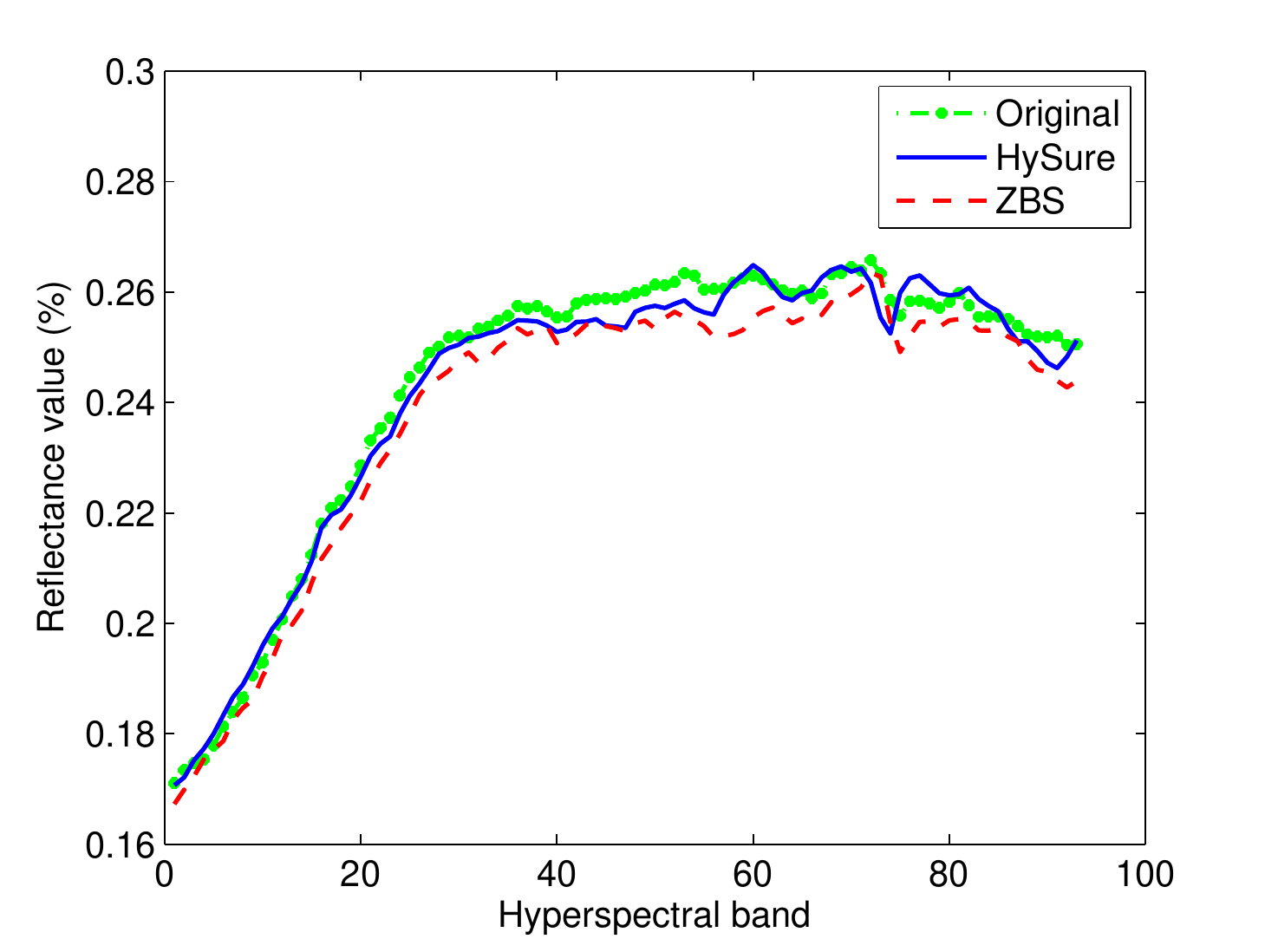}%
	\label{fig:rmse_error_plot10}}
\end{minipage}
	\begin{minipage}{0.33\linewidth}
				\centering
	\subfloat[{Reflectance of the pixel corresponding to the 50th percentile of the RMSE (dataset B).}]{\includegraphics[scale=.33]{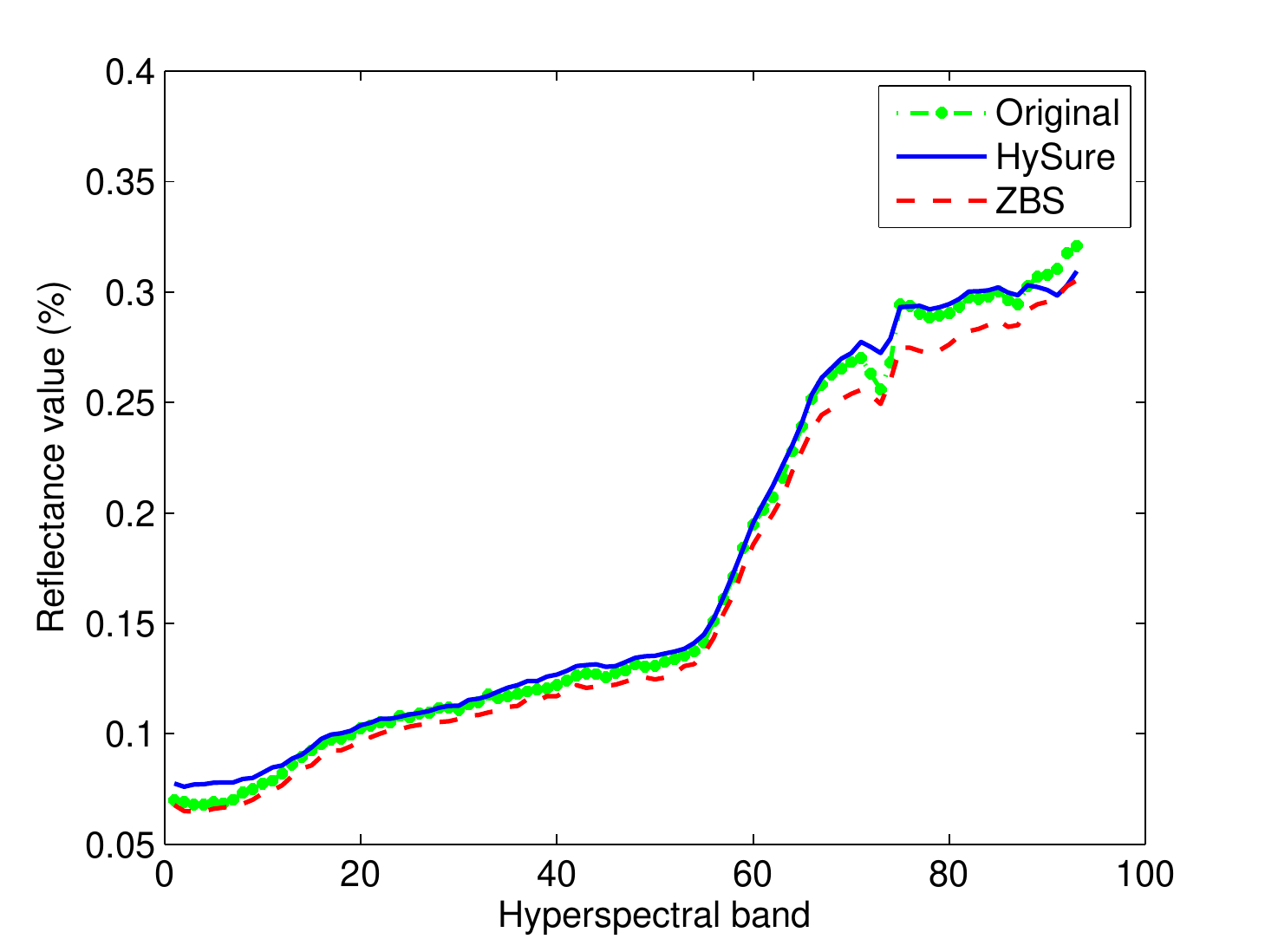}%
	\label{fig:rmse_error_plot50}}
\end{minipage}
	\begin{minipage}{0.32\linewidth}
				\centering
	\subfloat[{Reflectance of the pixel corresponding to the 90th percentile of the RMSE (dataset B).}]{\includegraphics[scale=.33]{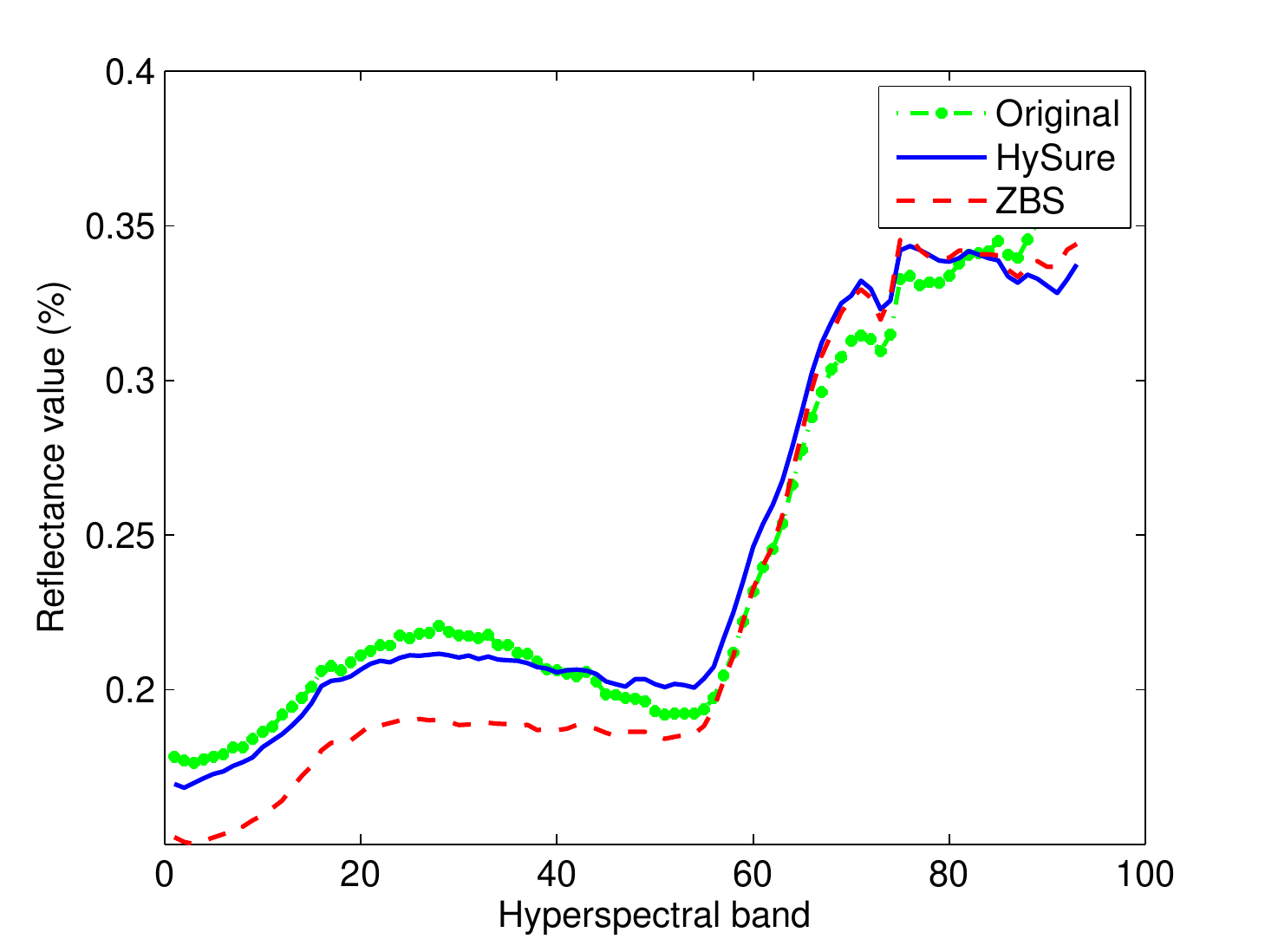}%
	\label{fig:rmse_error_plot90}}
\end{minipage}
	\caption{{Results for datasets B and C (HSI+MSI fusion). The results in Figs.{~\ref{fig:rmse_plot_hs_ms_sorted_pavia}} and~{\ref{fig:rmse_plot_hs_ms_sorted}} are in ascending order. Figs.~{\ref{fig:rmse_error_plot10}}, {\ref{fig:rmse_error_plot50}} and {\ref{fig:rmse_error_plot90}} show the reflectance values of the pixels corresponding to the 10th, 50th and 90th percentiles, respectively, for dataset B.}}
	\label{fig:rmse_error_plot}
\end{figure*}

The results of these tests are shown in Figs.~\ref{fig:pavia_ms} and~\ref{fig:paris_ms_hs} and in Tables \ref{tab:pavia_hs_ms} and~\ref{tab:paris_hs_ms}. The proposed method surpassed the other one in all tests. For dataset B, and due to input restrictions of the implementation of ZBS that was available to us, we only worked on a section of the image with $200 \times 200$ pixels, corresponding to the bottom left corner. For this dataset, as an illustration of the processing speed, the proposed method took about 35 seconds to perform the fusion in a MATLAB implementation running on an Intel\textsuperscript{\textregistered} Xeon\textsuperscript{\textregistered} CPU at 3.20~GHz with 16 GB of RAM. For dataset C, we worked on a section with $72 \times 72$ pixels. Fig.~\ref{fig:rmse_plot_hs_ms_sorted} shows the RMSE between the ground truth and the results of both methods, {for each pixel}, with the pixels sorted in order of ascending error; we are only showing results corresponding to the first 99\% of the errors, since the other pixels are very noisy. Figs.~{\ref{fig:rmse_error_plot10}}, {\ref{fig:rmse_error_plot50}} and {\ref{fig:rmse_error_plot90}} compare the reflectance values of the results of the two methods with the ground truth ones, for three pixels, corresponding to the 10th, 50th and 90th percentiles of the error of our method, respectively.

\begin{table}[!t]
\renewcommand{\arraystretch}{1.3}
\caption{Results for dataset B (HSI+MSI fusion).}
\label{tab:pavia_hs_ms}
\centering
\begin{tabular}{l||c|c|c}
\multicolumn{1}{r}{} & ERGAS & SAM & UIQI\\
\hline
\hline
ZBS & 5.919 & 4.375 & 0.881\\
\hline
\textbf{HySure} & \textbf{1.213} & \textbf{1.956} & \textbf{0.995}\\
\hline
\hline
\end{tabular}
\end{table}

\begin{table}[!t]
\renewcommand{\arraystretch}{1.3}
\caption{Results for dataset C (HSI+MSI fusion).}
\label{tab:paris_hs_ms}
\centering
\begin{tabular}{l||c|c|c}
\multicolumn{1}{r}{} & ERGAS & SAM & UIQI\\
\hline
\hline
ZBS & 5.011 & 3.672 & 0.725\\
\hline
\textbf{HySure} & \textbf{4.101} & \textbf{3.092} & \textbf{0.840}\\
\hline
\hline
\end{tabular}
\end{table}

\section{Conclusions} \label{sec:conclusions}

We have proposed a method, termed HySure, to perform the fusion of hyperspectral images with either panchromatic or multispectral ones, with the goal of obtaining images which have high resolution in both the spatial and the spectral domains. This problem is closely related to the pansharpening one, but presents new challenges due to the much larger size of hyperspectral images when compared with the multispectral images normally used in pansharpening and to the fact that the different images do not normally have a complete spectral overlap. In addition to performing the fusion, the proposed method is also able to estimate the relative spectral and spatial responses of the sensors from the data.

We formulated the fusion problem as a convex program, solved via the Split Augmented Lagrangian Shrinkage Algorithm (SALSA)---an instance of the Alternating Direction Method of Multipliers (ADMM). The estimation of the relative responses of the sensors was formulated as a convex quadratic program. Taking advantage of the low intrinsic dimensionality of hyperspectral images by working on a subspace of the space where those images are defined, and using an adequate variable splitting, we obtained an effective algorithm which compares quite favorably with several published methods on both simulated and real-life data.

\appendix  

%


In this Appendix we show in detail how to solve the optimization problem described in Section~\ref{sec:admm}. We start by expanding~(\ref{eq:al}) in its different components: 
\begin{equation} \label{eq:al2}
\begin{aligned}
\mathcal{L}\big(\X, &\V_1, \V_2, \V_3, \V_4, \A_1, \A_2, \A_3, \A_4 \big) \\
& \quad = \frac{1}{2}\Big\|\Y_h - \E\V_1\M \Big\|_F^2 + \frac{\mu}{2}\Big\|\X\B - \V_1 - \A_1\Big\|_F^2 \\ 
& \qquad + \frac{\lambda_{m}}{2}\Big\|\Y_m - \R\E\V_2 \Big\|_F^2 + \frac{\mu}{2}\Big\|\X - \V_2 - \A_2\Big\|_F^2 \\
& \qquad + \lambda_{\varphi} \varphi \big(\V_3, \V_4 \big) + \frac{\mu}{2}\Big\|\X\D_h - \V_3 - \A_3\Big\|_F^2 \\
& \qquad + \frac{\mu}{2}\Big\|\X\D_v - \V_4 - \A_4\Big\|_F^2~.
\end{aligned}
\end{equation}

The optimization algorithm, which was given in condensed form in Fig.~\ref{alg:salsa}, is given in more detail in Fig.~\ref{alg:salsa_details}.
\begin{figure}[tbh!]
\begin{center}
\colorbox{light}{\parbox{1.0\columnwidth}{
\begin{algorithmic}
\STATE $k \coloneqq 0$
\REPEAT
\STATE {$
\begin{aligned}
\mathbf{X}^{(k+1)} \in &\underset{\mathbf{X}}\argmin \mathcal{L}\Big(\mathbf{X}, \V_1^{(k)}, \cdots, \V_4^{(k)}, \\
& \quad \A_1^{(k)}, \cdots, \A_4^{(k)}\Big)
\end{aligned}
$
}
\vspace{6pt}
\FOR{$i = 1, \cdots, 4$}
\STATE {
$
\begin{aligned}
\mathbf{V}_i^{(k+1)} \in &\underset{\V_i}\argmin \mathcal{L}\Big(\mathbf{X}^{(k+1)}, \V_1^{(k)}, \cdots, \\
& \V_i^{(k)}, \cdots, \V_4^{(k)}, \A_1^{(k)}, \cdots, \A_4^{(k)}\Big)
\end{aligned}
$
}
\ENDFOR
\vspace{6pt}
\STATE {$\mathbf{A}_1^{(k+1)} \coloneqq \mathbf{A}_1^{(k)} - \Big( \mathbf{X}^{(k+1)} \B - \V_1^{(k+1)}$ \Big)}
\STATE {$\mathbf{A}_2^{(k+1)} \coloneqq \mathbf{A}_2^{(k)} - \Big( \mathbf{X}^{(k+1)} - \V_2^{(k+1)}$ \Big)}
\STATE {$\mathbf{A}_3^{(k+1)} \coloneqq \mathbf{A}_3^{(k)} - \Big( \mathbf{X}^{(k+1)} \D_h - \V_3^{(k+1)}$ \Big)}
\STATE {$\mathbf{A}_4^{(k+1)} \coloneqq \mathbf{A}_4^{(k)} - \Big( \mathbf{X}^{(k+1)} \D_v - \V_4^{(k+1)}$ \Big)}
\STATE {$k \coloneqq k + 1$}
\UNTIL{stopping criterion is satisfied.}
\end{algorithmic}
}}
\caption{Optimization algorithm.}
\label{alg:salsa_details}
\end{center}
\end{figure}

The first minimization problem is
\[
\begin{aligned}
\X^{(k+1)} \in \underset{\X}\argmin \frac{\mu}{2}&\Big\|\X \B - \V_1^{(k)} - \A_1^{(k)}\Big\|_F^2\\
\quad + \frac{\mu}{2}&\Big\|\X - \V_2^{(k)} - \A_2^{(k)}\Big\|_F^2\\
\quad + \frac{\mu}{2}&\Big\|\X \D_h - \V_3^{(k)} - \A_2^{(k)}\Big\|_F^2\\
\quad + \frac{\mu}{2}&\Big\|\X \D_v - \V_4^{(k)} - \A_3^{(k)}\Big\|_F^2,\\
\end{aligned}
\]
which has the solution
\begin{equation}
\begin{aligned}
\mathbf{X}^{(k+1)} = &\Big[\B \B^T + \eye + \D_h \D^T_h + \D_v \D_v^T \Big]^{-1}\\ 
&\Big[ \Big(\mathbf{V}_1^{(k)} + \A^{(k)}_1 \Big) \B^T + \Big(\mathbf{V}_2^{(k)} + \A^{(k)}_2 \Big)\\
&+ \Big(\mathbf{V}_3^{(k)} +\A_3^{(k)}\Big) \D_h^T + \Big(\mathbf{V}_4^{(k)} + \A_4^{(k)}\Big)\D_v^T\Big].\\[1mm]
\end{aligned}
\end{equation}
The computation can be efficiently performed through the use of the Fast Fourier Transform, having complexity $\mathcal{O}(L_s \times n_m \log n_m)$. The first term on the right hand side, including the inverse, can be computed in advance, before the iteration.

To solve the minimization problem involving $\V_1$,
\[
\begin{aligned}
& \mathbf{V}_1^{(k+1)} \in \underset{\mathbf{V}_1}{\argmin} \frac{1}{2}\Big\|\mathbf{Y}_h - \mathbf{EV}_1\mathbf{M} \Big\|_F^2\\
& \qquad + \frac{\mu}{2}\Big\|\mathbf{X}^{(k+1)}\B - \mathbf{V}_1 - \A_1^{(k)}\Big\|_F^2,
\end{aligned}
\]
we can take advantage of the masking matrix $\mathbf{M}$ to separate $\mathbf{V}_1$ into $\mathbf{V}_1\mathbf{M}$ and $\mathbf{V}_1 \overline{\mathbf{M}}$, where $\overline{\mathbf{M}}$ is the matrix that selects the pixels not selected by $\mathbf{M}$. We then have
\begin{equation}
\begin{aligned}
\mathbf{V}_1^{(k+1)}\mathbf{M} =&	\Big[\mathbf{E}^T\mathbf{E} + \mu \mathbf{I}\Big]^{-1}\\
& \quad \Big[\mathbf{E}^T\mathbf{Y}_h + \mu \Big(\mathbf{X}^{(k+1)}\B - \A_1^{(k)}\Big)\Big]\mathbf{M}
\end{aligned}
\end{equation}
and
\begin{equation}
		\mathbf{V}_1^{(k+1)}\overline{\mathbf{M}} = \Big(\mathbf{X}^{(k+1)}\B - \A_1^{(k)}\Big)\overline{\mathbf{M}}.
\end{equation}
$\left[\mathbf{E}^T\mathbf{E} + \mu \mathbf{I}\right]^{-1}$ and $\mathbf{E}^T\mathbf{Y}_h$ can be precomputed. The computations can be efficiently via the FFT, and have complexity $\mathcal{O}(L_s \times n_m \log n_m)$.

The minimization
\[
\begin{aligned}
 \mathbf{V}_2^{(k+1)} \in \underset{\mathbf{V}_2}{\argmin} \frac{\lambda_{m}}{2}&\Big\|\Y_m - \R\E\V_2 \Big\|_F^2 \\
 + \frac{\mu}{2}&\Big\|\X^{(k+1)} - \V_2 - \A_2^{(k)}\Big\|_F^2
\end{aligned}
\]
has the solution
\begin{equation}
\begin{aligned}
\mathbf{V}_2^{(k+1)} =&	\Big[\lambda_{m} \E^T \R^T \R \E + \mu \mathbf{I}\Big]^{-1}\\
& \quad \Big[\lambda_{m} \mathbf{E}^T \R^T \Y_m + \mu \Big(\mathbf{X}^{(k+1)} - \A_2^{(k)}\Big)\Big],
\end{aligned}
\end{equation}
where only $\left(\mathbf{X}^{(k+1)} - \A_2^{(k)}\right)$ cannot be precomputed. The complexity of this part is $\mathcal{O}(L_s \times n_m)$.

$\V_3$ and $\V_4$ are computed by solving the minimization problem
\[
\begin{aligned}
\Big\{\mathbf{V}_3^{(k+1)}, \mathbf{V}_4^{(k+1)}\Big\} &\in \underset{\mathbf{V}_3, \mathbf{V}_4}\argmin \lambda_{\varphi} \varphi \big(\V_3, \V_4 \big)\\
+ \frac{\mu}{2}&\Big\|\mathbf{X}^{(k+1)}\D_h - \mathbf{V}_3 - \A_2^{(k)}\Big\|_F^2 \\
+ \frac{\mu}{2}&\Big\|\mathbf{X}^{(k+1)}\D_v - \mathbf{V}_3 - \A_3^{(k)}\Big\|_F^2,\\
\end{aligned}
\]
whose solution is given by a column-wise \textit{vector-soft threshold} function~\cite{Donoho1995},
\begin{equation}
\begin{aligned}
& \Big\{\Big(\mathbf{V}_3^{(k+1)}\Big)_{: j}, \Big(\mathbf{V}_4^{(k+1)}\Big)_{: j} \Big\} = \\
& \qquad = \max \Big\{\big\|\mathbf{C}\big\|_F - \frac{\lambda_{\varphi}}{\mu}, 0 \Big\}\frac{\mathbf{C}}{\big\|\mathbf{C}\big\|_F},
\end{aligned}
\end{equation}
where
\[\mathbf{C}=\Big\{\Big(\mathbf{X}^{(k+1)}\D_h - \mathbf{A}_3^{(k)}\Big)_{: j}, \Big(\mathbf{X}^{(k+1)}\D_v - \mathbf{A}_4^{(k)}\Big)_{: j} \Big\},\]
and $(.)_{: j}$ denotes the $j$th column of a matrix. We follow the convention that $\mathbf{0} / ||\mathbf{0}||_F = \mathbf{0}$. The complexity of computing $\V_3$ and $\V_4$ is $\mathcal{O}(L_s \times n_m \log n_m)$, being dominated by FFTs.

After performing these optimizations, the following equations are used to update the Lagrange multipliers:
\[
\begin{aligned}
\A_1^{(k+1)} &= \A_1^{(k)} - \Big(\mathbf{X}^{(k+1)}\B - \mathbf{V}_1^{(k+1)}\Big),\\
\A_2^{(k+1)} &= \A_2^{(k)} - \Big(\mathbf{X}^{(k+1)} - \mathbf{V}_2^{(k+1)}\Big),\\
\A_3^{(k+1)} &= \A_3^{(k)} - \Big(\mathbf{X}^{(k+1)}\D_h - \mathbf{V}_3^{(k+1)}\Big),\\
\A_4^{(k+1)} &= \A_4^{(k)} - \Big(\mathbf{X}^{(k+1)}\D_v - \mathbf{V}_4^{(k+1)}\Big).\\
\end{aligned}
\]

The complexity of the algorithm is dominated by the FFTs, and is $\mathcal{O}(L_s \times n_m \log n_m)$ per iteration.

\section*{Acknowledgment}

We gratefully acknowledge Prof. Paolo Gamba for providing dataset B, Dr. Giorgio Licciardi for providing dataset C, and Qi Wei for providing dataset D. We also gratefully acknowledge Dr. Yifan Zhang for providing the source code for~\cite{Zhang2009} and Dr. Gemine Vivone for providing the source code for the different pansharpening algorithms (GS, GSA, FIHS, PCA, BT an HPF)~\cite{Vivone2014}.

\ifCLASSOPTIONcaptionsoff
  \newpage
\fi



\bibliographystyle{IEEEtran}
\bibliography{IEEEabrv,refs}
\end{document}